\documentclass{article} %
\usepackage{times}
\usepackage{arxiv}
\usepackage{amsmath}
\usepackage{hyperref}
\usepackage{url}
\usepackage{listings}
\usepackage{subcaption}

\usepackage[utf8]{inputenc} %
\usepackage[T1]{fontenc}    %
\usepackage{booktabs}       %
\usepackage{amsfonts}       %
\usepackage{nicefrac}       %
\usepackage{microtype}      %
\usepackage{lipsum}		%
\usepackage{graphicx}
\usepackage{natbib}
\usepackage{doi}
\title{TreeBoN: Enhancing Inference-Time Alignment with Speculative Tree-Search and Best-of-N Sampling}
\date{}

\usepackage{macro_commands}
\usepackage{xcolor}
\newcommand{\jjpar}[1]{\left( #1 \right)}

\begin{document}

\maketitle
\vspace{-7em}
\begin{center}
Jiahao Qiu$^{*1}$, Yifu Lu$^{*1}$, Yifan Zeng$^2$, Jiacheng Guo$^1$, Jiayi Geng$^3$, Chenhao Zhu$^4$, Xinzhe Juan$^5$,\\ Ling Yang$^1$, Huazheng Wang$^2$, Kaixuan Huang$^1$, Yue Wu$^1$, Mengdi Wang$^1$
\end{center}

\begin{center}
\textsuperscript{1}Princeton University \quad
\textsuperscript{2}Oregon State University \quad
\textsuperscript{3}Carnegie Mellon University \\
\textsuperscript{4}Stanford University \quad
\textsuperscript{5}University of Michigan \quad

\end{center}

\vspace{1em}

\def\thefootnote{*}\footnotetext{These authors contributed equally to this work.}

\begin{abstract}
Inference-time alignment enhances the performance of large language models without requiring additional training or fine-tuning but presents challenges due to balancing computational efficiency with high-quality output. Best-of-N (BoN) sampling, as a simple yet powerful approach, generates multiple responses and selects the best one, achieving improved performance but with a high computational cost. We propose TreeBoN, a novel framework that integrates a speculative tree-search strategy into Best-of-N (BoN) Sampling. TreeBoN maintains a set of parent nodes, iteratively branching and pruning low-quality responses, thereby reducing computational overhead while maintaining high output quality. Our approach also leverages token-level rewards from Direct Preference Optimization (DPO) to guide tree expansion and prune low-quality paths. We evaluate TreeBoN using AlpacaFarm, HH-RLHF, UltraFeedback, GSM8K, and TutorEval datasets, demonstrating consistent improvements. Specifically, TreeBoN achieves the highest win rate of 65\% on TutorEval and around 60\% win rates across other different datasets, outperforming standard BoN with the same computational cost and showcasing its scalability and alignment efficacy.
\end{abstract}
\def\thefootnote{\arabic{footnote}}

\section{Introduction}
Aligning large language models (LLMs) with human values is essential for ensuring their outputs reflect human intentions and ethical standards. When data on human preferences is available, a pretrained LLM can be fine-tuned to align with these preferences. One popular approach for fine-tuning is Reinforcement Learning from Human Feedback (RLHF), where a reward model is trained on a human-labeled preference dataset, followed by reinforcement learning to fine-tune the LLM as a policy model~\citep{ouyang2022training}. Alternative methods such as Direct Preference Optimization~\citep{rafailov2024direct} and its variants~\citep{azar2024general, ethayarajh2024kto, meng2024simpo} enable direct alignment via fine-tuning using a contrastive loss, eliminating the need for a separate reward model. 

This paper focuses on optimizing inference-time alignment of large language models (LLMs). By leveraging inference-time search, the capability of LLMs is enhanced during the generation process, improving real-time decision-making. Various techniques, such as Monte Carlo Tree Search (MCTS), have been effectively applied to reasoning, planning, and accelerated decoding tasks \citep{zhao2024large,hao2023reasoning,brandfonbrener2024verified,choi2023kcts}, demonstrating the potential for better decoding outcomes \citep{liu2023making}. 
In this work, we aim to explore tree search strategies to further capitalize on decoding-time alignment. Our goal is to enhance the quality of alignment while simultaneously reducing the computational cost of inference, providing a more efficient and aligned LLM experience.

\begin{figure*}
    \centering
    \includegraphics[width=0.9\linewidth]{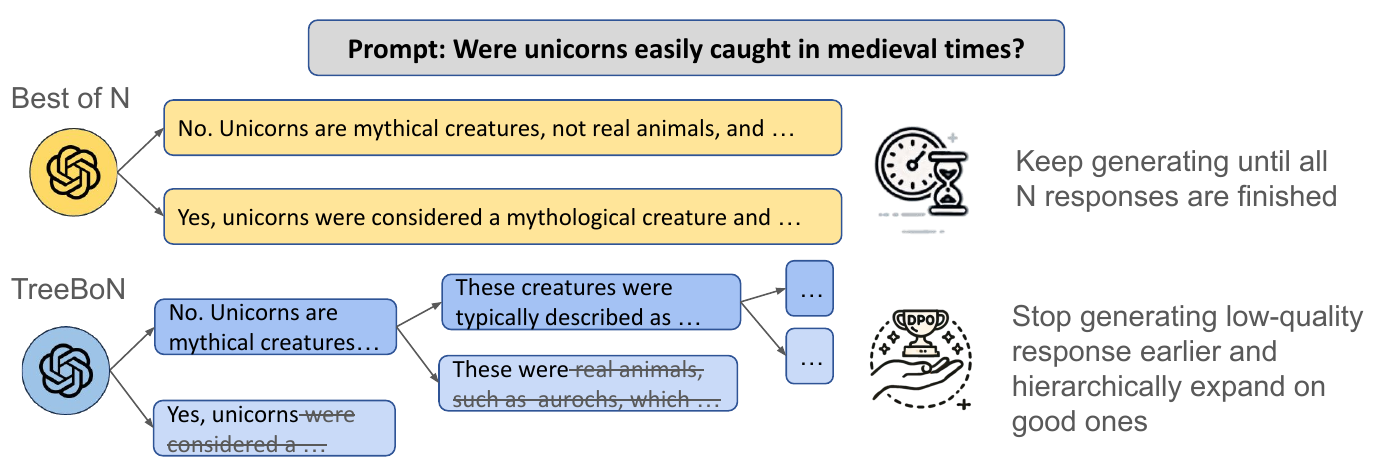}
    \caption{An illustration of different response generation strategies. Best-of-N completes all candidate generations, while TreeBoN (our method) introduces early termination of low-quality responses using a DPO reward model and hierarchically expands promising responses. See Table~\ref{tab:tok-reward-tree-response-detail} for the detailed example.}
    \label{fig:differnt-generation-strategies}
\end{figure*}

A most simple, yet powerful inference-time alignment method is the Best-of-N (BoN) method. We start our discussion with BoN to motivate our development of more efficient solutions.
BoN generates multiple sample responses and chooses the best one based on a reward function \( r(\yb | \xb) \) which characterizes how well-aligned a generated response \( \yb \) is with respect to the given prompt \( \xb \). More formally, BoN aims to approximate the solution to the following optimization problem:
\begin{align*}
    \max_{\yb} r(\yb | \xb),
\end{align*}    
where the only access to \( \yb \) is through auto-regressively sampling the next token \( y_t \) from the base policy \( \pi_{\text{base}}(\cdot| \xb, \yb_{1:t-1}) \), conditioned on the previous tokens. BoN generates $N$ samples and selects the response from \( \yb^{1}, \yb^{2}, \dots, \yb^{N} \) that achieves the highest reward model score. Due to its simplicity and effectiveness, Best-of-N sampling and its variants are widely studied to align LLM outputs with human preferences~\citep{wang2024improved, sessa2024bond, gui2024bonbon, khaki2024rs, jinnai2024regularized, liu2023statistical, xiong2024iterative}. Also, Best-of-N Sampling is commonly used in Expert Iteration and iterative fine-tuning~\citep{havrilla2024teaching}, which plays an important role in the alignment of Llama2~\citep{touvron2023llama} and Llama3~\citep{dubey2024llama}. In detail, Llama2~\citep{touvron2023llama} combines rejection sampling with Proximal Policy Optimization (PPO) in an iterative fine-tuning process to align Llama 2 with human preferences. More recently, Llama3~\citep{dubey2024llama} uses rejection sampling to generate high-quality data for alignment in an iterative process.   

While Best-of-N sampling has proven effective, it has a significant drawback: efficiency. Naively implementing BoN requires generating $N$ separate responses and the total inference FLOPs scales linearly with $N$. This not only demands $N$ times more computation but also potentially leads to $N$ times longer latency. The computational overhead can be prohibitively expensive for LLMs with billions of parameters, particularly when real-time or low-latency responses are needed. 

Some potential solutions involve more intelligent sampling strategies such as pruning to improve efficiency. \textit{Speculative Best-of-N} (SBoN)~\citep{zhang2024accelerating}, which aims to speed up the process while only sacrificing minimal performance on reward compared to the Best-of-N, alleviates the problem by continuing the generation of high-quality responses and rejecting the low-quality responses at an early stage of the generation. Cascade Reward Sampling(CARDS)~\citep{li2024cascade} use rejection sampling to iteratively generate small semantic segments to form such prefixes, based on rewards computed using incomplete sentences. 

These accelerated methods are based on the hypothesis that utterances receiving high/low rewards early on in the generation process are likely to yield high/low rewards in the final complete response. However, this hypothesis is too good to be true. In fact, off-the-shelf reward models are typically trained on complete responses, and therefore the score of partial completions by the reward model is usually chaotic and doesn't accurately predict the final output's quality, especially for long responses. Our analysis confirmed that rewards of partial completions are not necessarily positively correlated with the final reward (see our experiment results in Appendix \ref{appendix:rm_analysis}). 

To enable faster, efficient inference-time alignment, we propose to incorporate a tree search strategy into BoN sampling, in order to improve the alignment quality as well as reduce the overall inference cost. Our TreeBoN method maintains an active set of nodes, and actively grows a tree via branching and pruning. In other words, TreeBoN would sample more frequently from good parent nodes but prunes nodes with low predicted rewards. This tree search strategy makes it possible to efficiently explore the search space.

\begin{figure*}
    \centering
    \includegraphics[width=0.9\linewidth]{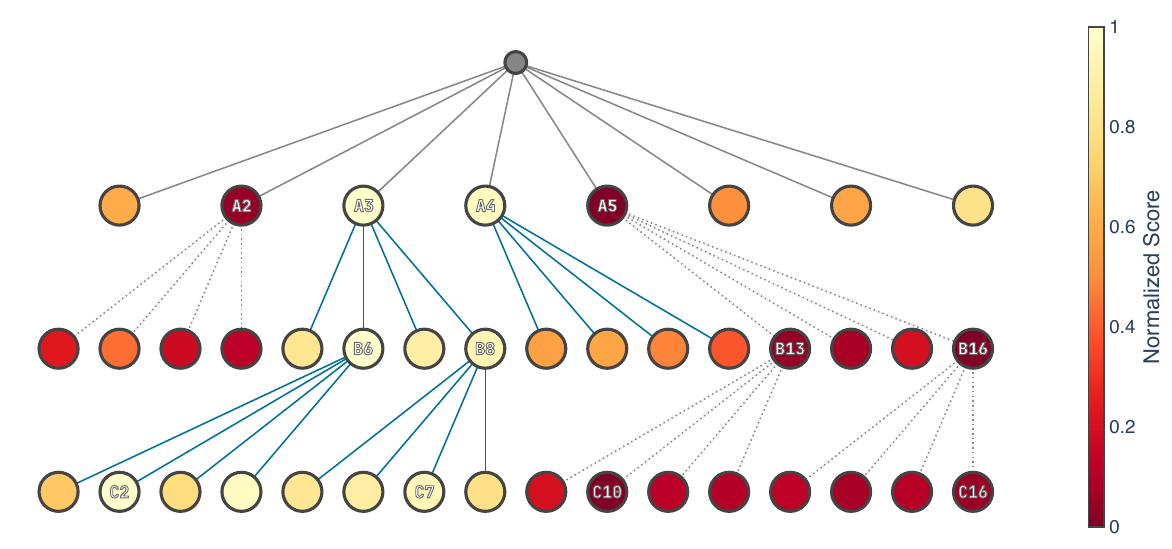}
    \caption{Visualization of speculative tree-search process for the prompt \textbf{"Were unicorns easily caught in medieval times?"}. Nodes represent partial responses, with color indicating normalized reward scores. We normalize the reward values within each layer. Solid blue lines show the expansion of high-reward paths, while dotted lines represent pruned low-reward branches. The solid blue line expansion path in this example shows the setting that $N=8$ initial candidate responses and $N_{\text{children}}=4$ in Algorithm \ref{alg:treebon}.
    In detail, labeled nodes \textbf{A2~(Yes, unicorns were considered a mythological creature and easily caught in medieval times)} and \textbf{A5~(Unicorns were believed to be easily caught in medieval times)} include hallucinations and therefore generating future responses from low-quality prefixes makes it hard to get a high-quality result.
    Meanwhile, \textbf{A3~(No. Unicorns are mythical creatures, not real animals, and therefore could not have been caught in medieval times)} and \textbf{A4~(No, unicorns were not easily caught in medieval times. In fact, unicorns were mythical creatures and did not exist in reality)} are high-reward prefixes that are more likely to produce high-quality complete responses in the future. More details of labeled nodes are presented in \textbf{Table \ref{tab:tok-reward-tree-response-detail}}. 
    } 
    \label{fig:generated-tree-with-reward}
\end{figure*}

Another design feature of TreeBoN is the use of implicit reward from DPO-aligned models for guidance of the tree research. DPO~\citep{rafailov2024direct} states that the DPO policy model can provide an implicit reward. \citet{rafailov2024r} further points out that DPO training implicitly learns a token-level reward function. Thus, we design TreeBoN to be able to leverage any off-the-shelf DPO model for inference-time decoding of the target model. With extensive experiments, we find that a weighted combination of implicit DPO rewards would lead to superior, robust performance. Our observation is consistent with the fact that one can detect safety levels of the full response using the first few tokens \citet{qi2024safety}.

Our experiments show that under the same computing budget, TreeBoN achieves better performance than BoN extensively and stably. It achieves the highest win-rates of 65\% on TutorEval~\citep{chevalier2024language}, 63\% on AlpacaFarm~\citep{alpacafarm} (length 192 and 384), above 60\% across HH-RLHF~\citep{bai2022training} and UltraFeedback~\citep{cui2024ultrafeedbackboostinglanguagemodels}, and increases pass@1 solve rate on GSM8K~\citep{cobbe2021gsm8k} as well. By choosing a smaller N, TreeBoN can simultaneously enhance performance and efficiency. With only 6.3\% of the compute, TreeBoN still maintains a 55\% win-rate against BoN. On the other hand, SBoN can be viewed as a special example of our method with a two-layer tree whose children number is equal to one and BoN can be viewed as a two-layer tree with the children number equal to N. TreeBoN has the potential to further improve efficiency than expected by taking advantage of the key-value cache which is especially beneficial to the tree structure since the keys and values of parent tokens can be cached and shared by children. 

The main contributions of this paper are as follows:
\begin{enumerate}
    \item We incorporate the Speculative Tree-search framework into Best-of-N Sampling to enhance efficiency and alignment performance simultaneously.
    \item We apply weighted implicit reward from DPO to provide the partial reward, which replaces the traditional reward model. We also offer a comprehensive analysis of traditional reward models on partial responses.\footnote{See Appendix~\ref{appendix:rm_analysis} for our sentence-level and token-level experiments and examples}
    \item TreeBoN demonstrates robust improvements in alignment quality and efficiency in comprehensive evaluations.
\end{enumerate}

\section{Related Works}
\subsection{Best-of-N Sampling for Alignment}
Best-of-N (BoN) sampling is a commonly used strategy for aligning large language models with human preferences by selecting the best sample out of N candidates. At training time, \cite{amini2024variational} fine-tunes models by minimizing the KL divergence to approximate the BoN distribution, improving value alignment using variational BoN, which reduces the computational cost during inference. \cite{sessa2024bond,gui2024bonbon} further enhance alignment by distilling the BoN sampling behavior directly into the model during training, aiming to replicate the BoN distribution with a single sample at inference time. At inference time, \cite{zhang2024accelerating} speeds up BoN by stopping the generation of unlikely candidates, and \cite{khaki2024rs} combines rejection sampling with preference optimization to improve efficiency without sacrificing alignment performance. From a theoretical perspective, an initial estimate for the KL divergence between the BoN output policy and the base model was provided for small values of N \cite{coste2023reward}, \cite{gao2023scaling}, \cite{go2023compositional}, and this estimate was later improved to cover all values of N \cite{beirami2024theoretical}. It has also been shown that BoN and KL-regularized reinforcement learning methods achieve similar asymptotic expected rewards, with minimal KL deviation between them \cite{yang2024asymptotics}. Compared with the works mentioned above, our work utilizes a tree-structured search scheme / segment-wise beam search to accelerate best-of-N sampling by pruning the low-reward branches early. To terminate low-reward branches early, we utilize the implicit value function from a DPO policy.

\subsection{Tree-Search/MCTS For Language Model}
MCTS has been employed in large language model tasks recently~\citep{kocsis2006bandit}. \citet{zhao2024large} and \citet{hao2023reasoning} integrates MCTS into planning and logical reasoning tasks. VerMCTS~\citep{brandfonbrener2024verified} utilizes a logical verifier to guide a modified Monte Carlo Tree Search (MCTS) for code generation. KCTS~\citep{choi2023kcts} guides the language model to generate text aligned with the reference knowledge at each decoding step by combining a knowledge classifier score and MCTS. PPO-MCTS\citep{liu2023making} combines MCTS and PPO value network for decoding. 

Speculative Decoding is introduced to accelerate LLM inference while keeping the distribution of LLM's output distribution unchanged by using a much smaller draft model to predict the LLM outputs which are verified later in parallel by the LLM~\citep{chen2023accelerating, leviathan2023fast}. SpecDec++~\citep{huang2024specdecboostingspeculativedecoding} adaptively selects candidate token lengths using a trained acceptance prediction head, achieving substantial inference speedups on large language models by reducing verification costs without sacrificing accuracy. SpecInfer and SpecTr extend the sequence to a token tree, increasing the number of accepted tokens by the target model~\citep{sun2024spectr, miao2023specinfer}. SEQUOIA further proposes the method for constructing the optimal tree structure for the speculated tokens by introducing a dynamic programming algorithm~\citep{chen2024sequoia}. Medusa~\citep{cai2024medusasimplellminference} is designed to accelerate large language model (LLM) inference by using multiple parallel decoding heads to predict multiple tokens simultaneously, reducing decoding steps without requiring a separate draft model, thus improving efficiency and speed while maintaining output quality.

While our tree-structured search framework bears resemblance with MCTS or tree-based speculative decoding, they are fundamentally different: most MCTS algorithms are designed for planning and logical reasoning tasks with a clear reward signal in the end, while our work focuses on using tree search to accelerate best-of-N sampling and LLM alignment and the signal is obtained throughout the search process. Tree-based speculative decoding is used to accelerate sampling from the target distribution, while ours is used to accelerate sampling for the best of the $N$ responses. PPO-MCTS doesn't consider the efficiency, instead, it focuses on token-level tree expansion involving the backup stage which takes more time. Also, the guidance of PPO-MCTS is a value network from PPO which differs from ours.

\subsection{Reward Modeling}
\textbf{Full-sequence reward modeling.} RLHF uses the Bradley-terry model to learn a reward function for full-sequence~\citep{christiano2017deep, stiennon2020learning}. DPO~\citep{rafailov2024direct} implicitly solves the KL-regularized RLHF problem by representing the reward with a language model.SimPO~\citep{meng2024simpo} considers a different BT model based on the average (length-normalized) reward rather than the sum of rewards. It is worth noting that alignment can go beyond a reward model due to the inconsistency in human preference. To this end \citep{ azar2023general, rosset2024direct, wu2024self}, also optimize LLM’s log-ratio according to different criteria, and the log-ratio can serve as sequence-level reward indicator. 

\textbf{Partial/Token-level reward modeling.} Not every token contributes to human preference equally. A token-level reward signal is thus desirable so that we can do credit assignments to each token. Reward grounding~\citep{yang2024preference} attempts to learn a token-level reward via Maximum Likelihood Estimation (MLE). They define a specific aggregation function so that token rewards can be transformed into sequence rewards, which can then be learned via MLE under the BT model.  Reward reshaping can also be used to obtain token-level rewards. For instance, \citet{chan2024dense} uses attention weights to redistribute the sequence reward to each token. \citet{mudgal2023controlled} and \citet{han2024value} propose learning a value function to guide token-level sampling in controlled decoding tasks.

\textbf{Inverse Q preference learning: DPO reward is a token-level reward model} More recent works go beyond reward modeling by treating the problem as inverse Q-learning. \citet{rafailov2024r} shows that the DPO loss can be interpreted as implicitly learning a token-level $Q^*$ function, represented by the LLM's logits. Similarly, Contrastive Preference Learning (CPL)~\citep{hejna2023contrastive} assumes that human preferences follow a Bradley-Terry model based on the sum of $Q$ values rather than the sum of rewards, and proposes to learn the $Q$ function directly. \citet{zeng2024token} similarly expand on this idea, presenting token-level direct preference optimization based on the Q value function.

In this work, we examine the effectiveness of these reward modeling approaches by incorporating these signals with our tree-search BoN framework. Additionally, we propose a new design: the weighted sum of implicit DPO rewards that turns out highly effective.

\subsection{Decoding-Time Alignment}
DeAL views decoding as a heuristic-guided search process and integrates alignment to decoding using a wide range of alignment objectives~\citep{huang2024deal}. RAD~\citep{deng2023reward} uses a unidirectional reward model and ARGS designs a weighted scoring function involving the reward model~\citep{khanov2024args} to do the reward-guided search for decoding-time alignment. URIAL~\citep{lin2023unlocking} and RAIN~\citep{li2023rain} use in-context learning by prompting the LLMs to do the self-alignment without SFT or RLHF. Controlled decoding ~\citep{mudgal2023controlled} trains a value function from the reward model for better token-level scoring. RLMEC~\citep{chen2024improving} trains a generative token-level reward model for alignment. Cascade Reward Sampling(CARDS)~\citep{li2024cascade} uses a reward model on semantically complete segments to accelerate the decoding. \citet{shi2024decoding} extends decoding-time alignment to multiple objectives by generating the next token from a linear combination of predictions of all base models.

Cascade Reward Sampling(CARDS)~\citep{li2024cascade} use rejection sampling to iteratively generate small semantically complete segments, based on rewards computed on incomplete responses by a reward model. The assumption is that semantically-complete high-reward prefixes induce high-reward complete text. However, as shown in Appendix~\ref{appendix:tokenrmanalysis}, for responses that are longer than 128 which are not included by CARDS, we show that in our tree search setting where partial responses are 1/3 of the length, the partial reward of a reward model, even on semantically complete segments, has little correlation to the reward on the full response, thus unsuitable to be combined with Tree-Search.

\section{Preliminaries}
\subsection{Best-of-N sampling (BoN)}
To approximate the optimization problem of maximizing the reward function \( r(\yb | \xb) \) which measures how well a generated response \( \yb \) sampled from the base policy \( \pi_{\text{base}}(\cdot| \xb) \) aligns with respect to the given prompt \( \xb \), Best-of-N Sampling (BoN) selects the response with the highest reward score from N independent and identically distributed (i.i.d.) responses generated by the language model $\pi_{\text{base}}$:
\begin{align*}
    \yb^{\star} = \argmax_{\yb \in \{\yb^{k}\sim \pi_{\text{base}}(\cdot | \xb)\}_{k=1}^{N} } r(\yb | \xb),
\end{align*}    
where the only access to \( \yb \) is through auto-regressively sampling the next token \( y_t \) from the base policy \( \pi_{\text{base}}(\cdot| \xb, \yb_{1:t-1}) \), conditioned on the previous tokens. 

\begin{algorithm}[H]
\caption{Best-of-N Sampling (BoN)}
\begin{algorithmic}[1]
    \STATE \textbf{Input:} Prompt $\xb$, base policy \( \pi_{\text{base}} \), reward model \( r \), number of samples $N$, max length $l_{\text{max}}$.
    \STATE \textbf{Output:} Response $\yb^{\star}$ with the highest reward using BoN.
    
    \STATE \textbf{Initialization:} Generate $N$ responses $ \{ \yb^1, \yb^2, \dots, \yb^N \}$, each with maximum length $l_{\text{max}}$.
    \STATE Query the reward model to compute the reward scores $ r(\yb | \xb) $ for each generated response $\yb \in  \{ \yb^1, \yb^2, \dots, \yb^N \}$.
    \STATE Find the response $\yb^{\star}$ with the highest reward:
    \[
    \yb^{\star} = \argmax_{\yb \in  \{ \yb^1, \yb^2, \dots, \yb^N \}} \ r(\yb | \xb).
    \]
    
    \STATE \textbf{Return} the response $\yb^{\star}$.
\end{algorithmic}
\end{algorithm}

\subsection{Token-Level Markov Decision Process and Soft Q-Learning}

\citet{rafailov2024direct} demonstrated that under the Max-Entropy reinforcement learning (RL) formulation, the token-level log-ratio can be interpreted as an implicit token-level reward or advantage function, which remains invariant under reward shaping. 
Below, we briefly restate the key setting and results.

The token-level Markov Decision Process (MDP) defines the state $\mathbf{s}_t = (x_1, x_2, \dots, x_m, y_1, y_2, \dots, y_t)$ as the tokens generated so far, and the action $\mathbf{a}_t = y_{t+1}$ as the next token to be predicted. The auto-regressive language model is thus a policy $\pi(\mathbf{a}_t|\mathbf{s}_t)$. 
The transition dynamics are deterministic: $\sbb_{t+1} = \sbb_t | \ab_t$, simply appending the next token to the current generated tokens to form a new sequence. 

The RLHF formulation is expressed as a reverse-KL regularized reward maximization problem:
\[
\max_{\btheta} \EE_{\xb \sim \mathcal{X}, \yb \sim \pi_{\btheta}(\cdot|\xb)} \left[ r(\yb | \xb) \right] - \beta \EE_{\xb \sim \mathcal{X}} \left[\mathrm{KL}(\pi_{\btheta}(\cdot|\xb) \|\pi_{\text{ref}}(\cdot|\xb))\right],
\]
which can be rewritten as a Max-Entropy RL problem:
\[
\EE_{\xb \sim \mathcal{X}, \yb \sim \pi_{\btheta}(\cdot|\xb)} \big[{r(\yb | \xb) + \beta \log \pi_{\text{ref}}(\yb|\xb) }  \big] + \beta \EE_{\xb \sim \mathcal{X}} \left[\mathcal{H}(\pi_{\btheta}(\cdot|\xb))\right].
\]
Or equivalently at the token level:
\[
\EE_{\sbb_0 \sim \mathcal{X}, \ab_t \sim \pi_{\btheta}(\cdot|\sbb_t)} \left[ \sum_{t=1}^{T} r'(\sbb_t,\ab_t) \right] + \beta \EE_{\sbb_0 \sim \mathcal{X}} \left[\mathcal{H}(\pi_{\btheta}(\cdot|\sbb_0))\right],
\]
with the token level reward function $r'$ defined as:
\begin{align*}
    r'(\sbb_t,\ab_t)
    & :=
    \begin{cases}
        \beta \log \pi_{\text{ref}}(\ab_t | \sbb_t),  \text{ if $\sbb_{t+1}$ is not terminal,} \\
        r(\yb | \xb) + \beta \log \pi_{\text{ref}}(\ab_t | \sbb_t), \text{ if $\sbb_{t+1} = (\xb,\yb)$ is terminal.} 
    \end{cases}
\end{align*}

For simplicity, let us assume that the horizon is fixed at $T$. The derivation of the Max-Entropy RL formulation~\citep{ziebart2010modeling,rafailov2024r} utilizes the (soft) optimal value function $V^*$ and the (soft) optimal $Q$-function $Q^*$, as follows:
\[
V^*(\mathbf{s}_{T+1}) = 0, \quad \text{(terminal state)},
\]
\[
Q^*(\mathbf{s}_t, \mathbf{a}_t) = r'(\sbb_t,\ab_t) + V^*(\mathbf{s}_{t+1}),
\]
\[
V^*(\mathbf{s}_t) = \log \sum_{\mathbf{a}} \exp(Q^*(\mathbf{s}_t, \mathbf{a})), \quad \text{when } t \leq T.
\]
The optimal policy $\pi^*$ satisfies the following equation:
\[
\beta \log \pi^*(\mathbf{a}_t | \mathbf{s}_t) = Q^*(\mathbf{s}_t, \mathbf{a}_t) - V^*(\mathbf{s}_t),
\]
which can be further rewritten when $t < T$:
\[
\beta \log \frac{\pi^*(\mathbf{a}_t | \mathbf{s}_t)}{\pi_{\text{ref}}(\mathbf{a}_t | \mathbf{s}_t)} = V^*(\mathbf{s}_{t+1}) - V^*(\mathbf{s}_t).
\]
This suggests that we can use the partial sum of the implicit reward from a DPO policy to characterize the potential final reward given a prefix sequence of length $K$:
\begin{align*}
\sum_{t=0}^{K-1}    
\beta \log \frac{\pi^*(\mathbf{a}_k | \mathbf{s}_k)}{\pi_{\text{ref}}(\mathbf{a}_k | \mathbf{s}_k)} = V^*(\mathbf{s}_{K}) - V^*(\mathbf{s}_0).   
\end{align*}
Since $\sbb_0 = (x_1, x_2, \dots, x_m) = \xb$, $V^*(\mathbf{s}_0)$ is the same for all responses.

\section{Method}
\label{sec:method}
In this section, we introduce \textbf{TreeBoN}, a novel inference-time algorithm that enhances alignment quality and efficiency by incorporating a speculative tree-search structure into the Best-of-N (BoN) sampling framework. TreeBoN iteratively expands high-reward partial responses, pruning low-quality candidates at early stages. The algorithm leverages a weighted implicit reward from a Direct Preference Optimization (DPO) policy model to improve the quality of partial response evaluation. Below, we describe the key steps involved in TreeBoN.

\subsection{Overview of TreeBoN Algorithm}

TreeBoN operates by generating candidate responses layer-by-layer in a tree structure. The algorithm begins with a set of initial root responses, and at each subsequent layer, only high-reward responses are selected and expanded into multiple children. This speculative search through the tree space improves both the efficiency and the final response quality. The overall structure of TreeBoN is illustrated in Algorithm~\ref{alg:treebon} and Figure~\ref{fig:generated-tree-with-reward}. 

The algorithm takes as input the prompt $\xb$, a base policy $\pi_{\text{base}}$ for generating candidate responses, a partial-reward function \(r\), and key hyper-parameters including the number of root samples \(N\), maximum response length \(l_{\text{max}}\), branching factor(number of children per node) \(N_{\text{children}}\), and the number of tree layers \(N_{\text{layer}}\). 

Furthermore, $C_i$ denotes the candidate set containing all partial responses generated in the i-th layer. $P_i$ denotes the i-th layer active set containing all promising partial responses for expansion in the next layer. $l_i$ is the max new token length for generation in each layer, where $l_i = \frac{l_{\text{max}}}{N_{\text{layer}}}$.

\begin{algorithm}[H]
\caption{TreeBoN Algorithm}\label{alg:treebon}
\begin{algorithmic}[1]
    \STATE \textbf{Input:} Prompt $\xb$, base policy \( \pi_{\text{base}} \), partial-reward function \( r \), number of root samples $N$, max length $l_{\text{max}}$, branching factor $N_{\text{children}}$, number of tree layers $N_{\text{layer}}$.
    \STATE \textbf{Output:} Response $\yb^{\star}$ with the highest reward using TreeBoN.
    
    \STATE \textbf{Initialization:} Split the total max length $l_{\text{max}}$ into segments $l_1, l_2, \dots, l_{N_{\text{layer}}}$ where $l_i = \frac{l_{\text{max}}}{N_{\text{layer}}}$.
    \STATE Generate $N$ initial candidate responses for the first-layer candidate set $C_1 = \{\yb^1, \yb^2, \dots, \yb^N\}$, each with a length of $l_1$.

    \FOR{$i=1$ \TO $N_{\text{layer}}-1$}
        \STATE Query the reward model or partial reward function \( r(\yb | \xb) \) to compute the reward scores for each candidate response $\yb \in C_i$.
        \STATE Select the top $\frac{N}{N_{\text{children}}}$ candidates from $C_i$ based on reward scores to form the i-th layer active set $P_i$.
        \FOR{each parent response $\yb \in P_i$}
            \STATE For each parent $\yb$, continue generation by sampling $N_{\text{children}}$ child responses from the base policy \( \pi_{\text{base}} \), each with a max new token length $l_{i+1}$, to form the next set of candidates $C_{i+1}$.
        \ENDFOR
    \ENDFOR
    
    \STATE After all layers are generated, query the reward model for the final set of responses $C_{N_{\text{layer}}}$.
    \STATE Find the response $\yb^{\star}$ with the highest reward:
    \[
    \yb^{\star} = \argmax_{\yb \in C_{N_{\text{layer}}}} r(\yb | \xb).
    \]
    
    \STATE \textbf{Return} the response $\yb^{\star}$.

\end{algorithmic}
\end{algorithm}

\subsection{TreeBoN Generation Process}

The generation process in TreeBoN consists of the following key steps:

\begin{enumerate}
    \item \textbf{Initial Candidate Generation:} TreeBoN begins by generating $N$ candidate responses \(C_1 = \{\yb^1, \yb^2, \dots, \yb^N\}\) with a length of $l_1$ using the base policy \(\pi_{\text{base}}\). The total maximum response length \(l_{\text{max}}\) is split into segments \(l_1, l_2, \dots, l_{N_{\text{layer}}}\) evenly where $l_i = \frac{l_{\text{max}}}{N_{\text{layer}}}$.
    
    \item \textbf{Partial Reward Scoring:} At each layer \(i\), the reward model or partial-reward function \(r(\yb | \xb)\) is used to compute the reward score for each candidate response \(\yb \in C_i\). This is performed after generating partial responses of length \(l_i\).
    
    \item \textbf{Pruning and Selection:} Based on the reward scores, the top $\frac{N}{N_{\text{children}}}$ candidates from the current layer are selected to form the active set \(P_i\). These high-reward parent responses are used to generate child responses at the next layer.
    
    \item \textbf{Response Expansion:} For each parent response \(\yb \in P_i\), TreeBoN generates $N_{\text{children}}$ child responses by sampling from the base policy \(\pi_{\text{base}}\) with a maximum new token length \(l_{i+1}\). This process generates the next-layer candidate set \(C_{i+1}\). It is worth noting that the set size of the candidate set is always $N$ and the set size of $P_i$ is always $\frac{N}{N_{\text{children}}}$ to ensure an equal number of total generated tokens without requiring extra computing budget.
    
    \item \textbf{Final Selection:} After generating candidates for all layers, the reward model computes the final rewards for the candidate responses in the last layer \(C_{N_{\text{layer}}}\). The response \(\yb^{\star}\) with the highest reward is selected as the final output:
    \[
    \yb^{\star} = \argmax_{\yb \in C_{N_{\text{layer}}}} r(\yb | \xb).
    \]
\end{enumerate}

\subsection{Weighted Implicit Reward As Guidance}

One of the key contributions of TreeBoN is the use of a weighted implicit reward function, inspired by \citet{rafailov2024direct, rafailov2024r, qi2024safety}, to evaluate partial responses. This approach allows TreeBoN to replace the traditional reward model with a DPO policy model, which provides more accurate rewards for incomplete responses. The partial reward for a sequence \(\yb_{:K}\) is computed as:

\begin{align*}
    r_{\text{partial}}(\yb_{:K} | \xb)
    & =
    \sum_{k=0}^{K-1} w_k \log \frac{\pi^*(y_k | \xb, \yb_{:k})}{\pi(y_k | \xb, \yb_{:k})},
\end{align*}
where \(w_k = \frac{1}{|\yb_{:k}|}\) acts as a weighting factor to adjust the contribution of each token-level log-likelihood ratio. This weighted reward helps prune low-quality responses early and encourages the continuation of higher-quality candidates throughout the tree expansion process. We also test several different variants of partial reward modeling in section \ref{sec:diff_reward}.

\subsection{Comparison to Baseline Methods}

TreeBoN builds upon and extends earlier sampling strategies, such as \textit{Accelerating Best-of-N via Speculative Rejection} (SBoN)~\citep{zhang2024accelerating}, by integrating a speculative tree-search framework and partial reward function. SBoN relies on the assumption that partial-reward scores are positively correlated with full-response rewards. However, this assumption often leads to suboptimal performance in alignment tasks due to the inaccurate scoring of partial responses by reward models which are typically trained on complete responses. TreeBoN addresses this limitation by utilizing a more precise implicit reward signal derived from the Direct Preference Optimization (DPO) policy model, which significantly enhances the reliability of partial-reward approximation.

Moreover, TreeBoN leverages a hierarchical tree structure to explore the response space more comprehensively, balancing both alignment quality and computational efficiency. This tree-based approach allows for more flexible and effective pruning of low-quality responses while expanding promising candidates over multiple layers. As a result, TreeBoN can be seen as a generalization of SBoN, where setting $N_{\text{children}} = 1$ and $N_{\text{layer}} = 2$ reduces TreeBoN to the two-layered structure of SBoN.

Compared to traditional Best-of-N (BoN) sampling, which explores candidate responses without any hierarchical structure, TreeBoN employs a more structured exploration strategy. By generating and refining responses layer by layer, TreeBoN achieves a more efficient search of the response space using fewer overall samples. This leads to improvements in both speed and performance, as the tree-based generation effectively balances the trade-off between exploration and exploitation.

TreeBoN can be further accelerated while maintaining high alignment quality by taking advantage of key-value caching mechanisms, particularly beneficial in the tree structure, where the keys and values of parent tokens can be reused by their children. 

\section{Experiments}

\subsection{Experiment Setting}
We use a set of Llama models: LLaMA3-iterative-DPO-final~\citep{xiong2024iterative, dong2024rlhf} as the DPO policy model (referred as the DPO model in this section)\footnote{See model card \url{https://huggingface.co/RLHFlow/LLaMA3-iterative-DPO-final}}, with its SFT (supervised fine-tuning) checkpoint trained from Llama 3 8B~\citep{llama3} and reward model 
FsfairX-LLaMA3-RM-v0.1~\citep{dong2023raft,xiong2024iterative} from Llama 3 8B Instruct~\citep{llama3}. The SFT model was trained on a set of high-quality
instruction datasets for 1 epoch; the reward model was formulated as a Bradley-Terry model optimizing the negative log-likelihood loss function on a mixture of filtered datasets; and notably, the DPO policy model was initialized from the SFT model and updated on the online preference signals produced by the aforementioned reward model (as a proxy of human feedback). We refer readers to~\cite{xiong2024iterative} for the details of iterative online RLHF and the training of these models. We also use an additional DPO model Llama-3-8B-SFR-Iterative-DPO-R\footnote{This is the official release, trained with same SFT and reward model, see model card for details \url{https://huggingface.co/Salesforce/LLaMA-3-8B-SFR-Iterative-DPO-R}}, referred as the SFR model in this section. The baseline is the Best-of-N sampling with $N$ equal to 128 and the max token length of responses varies from 192 to 768. For Tree-based BoN with Weighted Implicit Reward, unless otherwise specified, we set the number of tree layers as 4, the number of children per node 4. Considering the cost of the evaluation, we take 100 randomly selected samples from each dataset, following the same setting as SBoN~\citep{zhang2024accelerating}.
We evaluate the baseline and our methods and take the average of 3 runs of different seeds on AlpacaFarm~\citep{alpacafarm}, UltraFeedback~\citep{cui2024ultrafeedbackboostinglanguagemodels}, GSM8K~\citep{cobbe2021gsm8k}, and HH-RLHF \citep{bai2022training}. For TutorEval~\citep{chevalier2024language}, we choose 100 closed-book questions. We also conduct experiments with process reward model (PRM)~\citep{lightman2023letsverifystepstep} on reasoning task. Similarly, we use Llama models: LLaMA3-iterative-DPO-final and Llama-3-8B-SFR-Iterative-DPO-R, on GSM8K, and we randomly sample 100 questions in 3 runs of same seeds as before. We take the average of process rewards for each step as the final reward ~\citep{lightman2023letsverifystepstep} and choose the \texttt{Qwen2.5-Math-PRM-7B} \footnote{\url{https://huggingface.co/Qwen/Qwen2.5-Math-PRM-7B}} ~\citep{zhang2025lessonsdevelopingprocessreward} as the PRM. We experiment with different maximum token lengths, number of candidates, and generative models, applying both BoN and TreeBoN approaches.

\subsubsection{Metrics} 
\textbf{GPT4 Win-rate}
For all datasets except for GSM8k, we conduct the standard GPT4 win-rate evaluations of our proposed method against the baseline. More specifically, given the same prompt, a response from the baseline and a response from the compared method are fed to an automatic evaluator of AlpacaEval~\citep{alpaca_eval} with randomized positions, which then formats them into a prompt, and asks GPT4~\citep{achiam2023gpt} to rank both responses.\footnote{We use the default alpaca\_eval\_gpt4 automatic evaluator. See \url{https://github.com/tatsu-lab/alpaca_eval} for the prompt and other details.} 

\textbf{Pass@1 Solve Rate }
For GSM8k, we report the zero-shot pass@1 solve rate~\citep{cobbe2021gsm8k}. Pass@k measures the rate of successfully passing the test (answering the math question correctly) from the $k$ responses that the algorithm generates. Thus, pass@1 means that the algorithm only outputs one response per question.\footnote{Though both BoN and TreeBoN generate multiple responses, only the final one response picked by the algorithm is considered the output and evaluated.} We extract the number after "answer is " as the final answer.

\textbf{FLOPs}
We consider FLOPs as a cost metric, and the Llama models we are using are all based on Llama 3 8B, which has 8 billion parameters, 32 layers, a context length of 8192, and token dimensions of 4096. The cost of LLMs mainly arises from the number of generated tokens and the matrix multiplications for dense transformers like Llama 3, considering the practical implementations of KV Cache that enable keys and values of parent tokens to be reusable (for the reward model and DPO model as well), we can approximate inference FLOPs with the same formula as in~\cite{brown2024largelanguagemonkeysscaling}:
\begin{align*}
    \text{FLOPs per token} &\approx 2 * \jjpar{ \text{num parameters} + 2 * \text{num layers} * \text{token dim} * \text{context length}}\\
    &= 2 * \jjpar{ 8 * 10^9 + 2 * 32 * 4096 * 8192}\\
    &\approx 2 * 10^{10} .\\
    \text{total inference FLOPs for BoN} &\approx 2 * ( \text{num prompt tokens} * \text{FLOPs per token}\\ &+ l_{\text{max}} * N * \text{FLOPs per token} ).\\
     \text{total inference FLOPs for TreeBoN}&\approx 2 * ( \text{num prompt tokens} * \text{FLOPs per token}\\ &+ \frac{l_{\text{max}} }{N_{\text{layer}}} * N * \text{FLOPs per token}\\ &+ (N_{\text{layer}}-1) * \frac{l_{\text{max}}}{N_{\text{layer}}} * N_{\text{children}} * \frac{N}{N_{\text{children}}} * \text{FLOPs per token} ) \\ &= \text{total inference FLOPs for BoN.}
\end{align*}
The extra multiplication of a factor of 2 is due to the cost of running a reward model for BoN and a DPO model for TreeBoN. We can see that in our setup, the computation cost of TreeBoN and Best-of-N will only be controlled by the number of root samples $N$ and maximum generation length $l_{\text{max}}$.

\subsection{Results}

\begin{figure}[htbp]
    \centering
    \begin{minipage}{0.49\linewidth}
        \centering
        \includegraphics[width=0.9\linewidth]{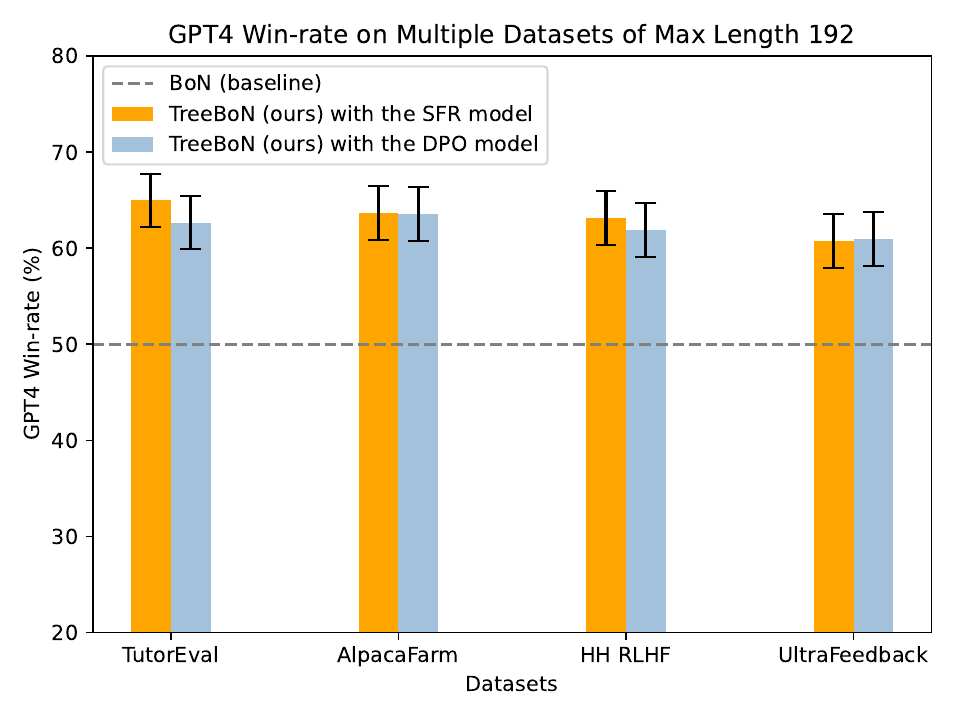}
        \caption*{Maximum length $192$.}
    \end{minipage}
    \begin{minipage}{0.49\linewidth}
        \centering
        \includegraphics[width=0.9\linewidth]{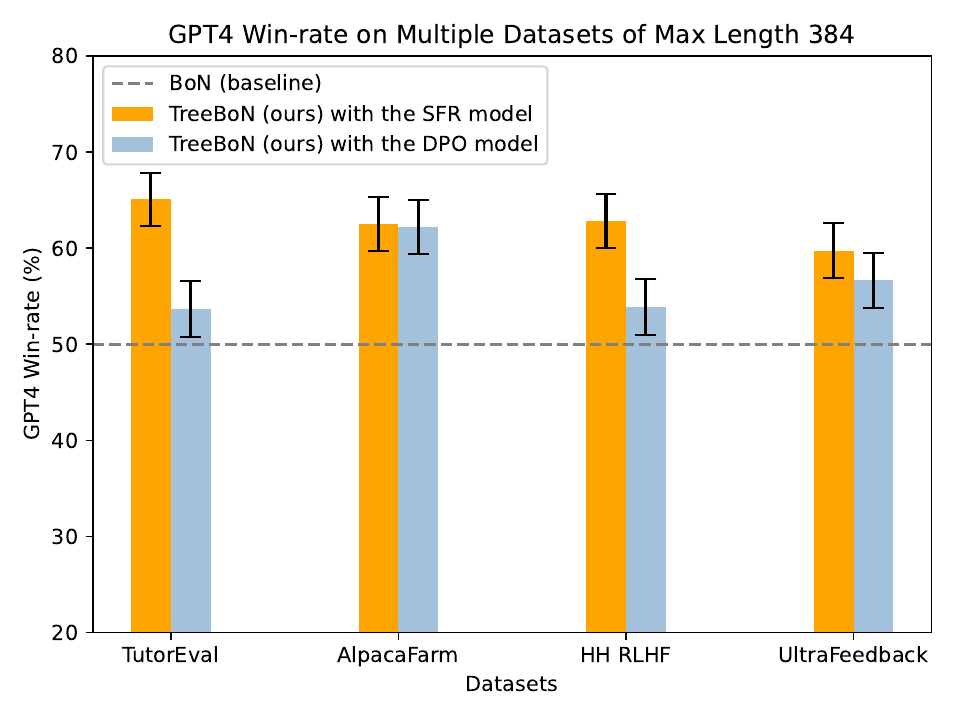}
        \caption*{Maximum length $384$.}
    \end{minipage}
    \caption{GPT4 win-rate of TreeBoN against BoN on multiple datasets. The SFR model refers to using Llama-3-8B-SFR-Iterative-DPO-R as the DPO model, and the DPO model refers to using LLaMA3-iterative-DPO-final.}
    \label{fig:mul_dataset}
\end{figure}

\begin{figure}
    \centering
    \includegraphics[width=0.5\linewidth]{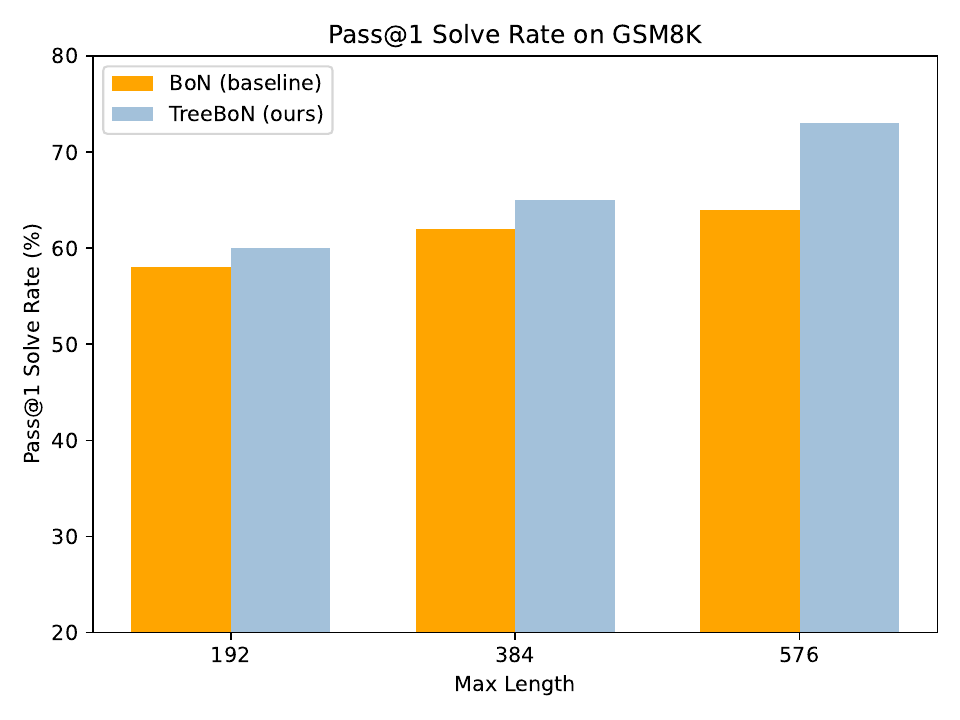}
    \caption{Pass@1 Solve Rate of TreeBoN with the DPO model and BoN on GSM8K.} 
    \label{fig:llama_gsm8k}
\end{figure}

\subsubsection{Improvement over Diverse Datasets}
We evaluate the baseline and our methods by answering 100 randomly selected prompts from AlpacaFarm~\citep{alpacafarm}, UltraFeedback~\citep{cui2024ultrafeedbackboostinglanguagemodels}, and HH-RLHF \citep{bai2022training}. For TutorEval~\citep{chevalier2024language}, we choose 100 closed-book questions. TreeBoN consistently outperforms the baseline across various datasets when evaluated using GPT-4 win-rate (Figure~\ref{fig:mul_dataset}). The full numerical results of this section can be found in Table~\ref{tab:sfr_mul_dataset},~\ref{tab:mul_dataset} and~\ref{tab:llama_gsm8k} of Appendix~\ref{appendix:res}.

Notably, with a maximum length of 192 tokens, TreeBoN with the SFR model achieves a 65\% win-rate than Best-of-N sampling on TutorEval, a 64\% win-rate on AlpacaFarm, and at least 60\% win-rate on other datasets. TreeBoN with the DPO model also achieves a 64\%win-rate on AlpacaFarn, and at least 60\% on others. This demonstrates that TreeBoN’s layered tree structure, combined with the use of a weighted implicit reward function to evaluate partial responses, enables better alignment with human preferences.

For longer responses (max length 384 tokens), TreeBoN with the SFR model maintains a significant performance lead, showing a 65\% win-rate over BoN on TutorEval, 63\% on AlpacaFarm and HH-RLHF. If using the DPO model, TreeBoN achieves a 62\% win-rate on AlpacaFarm as well. Notably, for the SFR model, from length 384 to length 768, the win-rates are steadily high. This suggests that TreeBoN is also well-suited for handling tasks that require generating more complex or nuanced responses, where multiple layers of exploration yield better results than repeated sampling.

In addition to general alignment improvements, TreeBoN’s zero-shot performance on mathematical reasoning dataset GSM8K~\citep{cobbe2021gsm8k} also sees a non-trivial boost. In Figure~\ref{fig:llama_gsm8k}, TreeBoN with the DPO model outperformed BoN by an impressive 9\% margin of pass@1 solve rate at maximum response lengths of 576 tokens, indicating that the hierarchical nature of TreeBoN allows it to effectively manage challenging reasoning tasks that require long CoT reasoning, making it adaptable across different domains.

\subsubsection{Explore Different Tree Structures with Same Computation}
\begin{figure}[h]
    \centering
    \begin{minipage}{0.49\linewidth}
        \centering
        \includegraphics[width=0.9\linewidth]{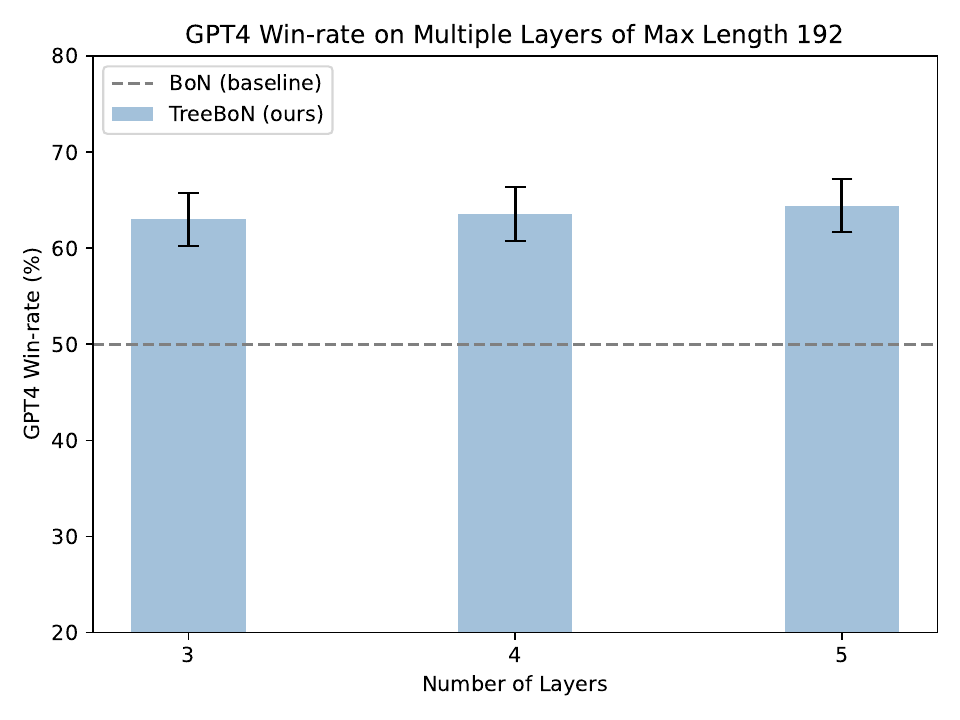}
         \caption*{Maximum length $192$.}
    \end{minipage}
    \begin{minipage}{0.49\linewidth}
        \centering
        \includegraphics[width=0.9\linewidth]{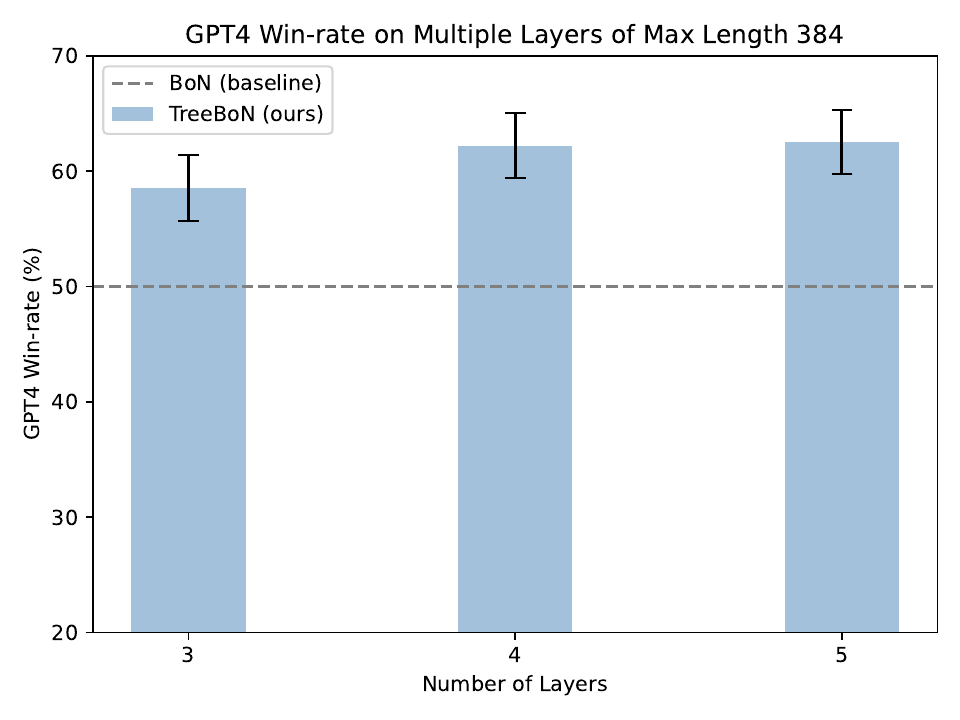}
        \caption*{Maximum length $384$.}
    \end{minipage}
    \caption{GPT4 win-rate of TreeBoN against BoN on AlpacaFarm with the different number of tree layers.}
    \label{fig:ablate_layer}
\end{figure}

\begin{figure}[h]
    \centering
    \begin{minipage}{0.49\linewidth}
        \centering
        \includegraphics[width=0.9\linewidth]{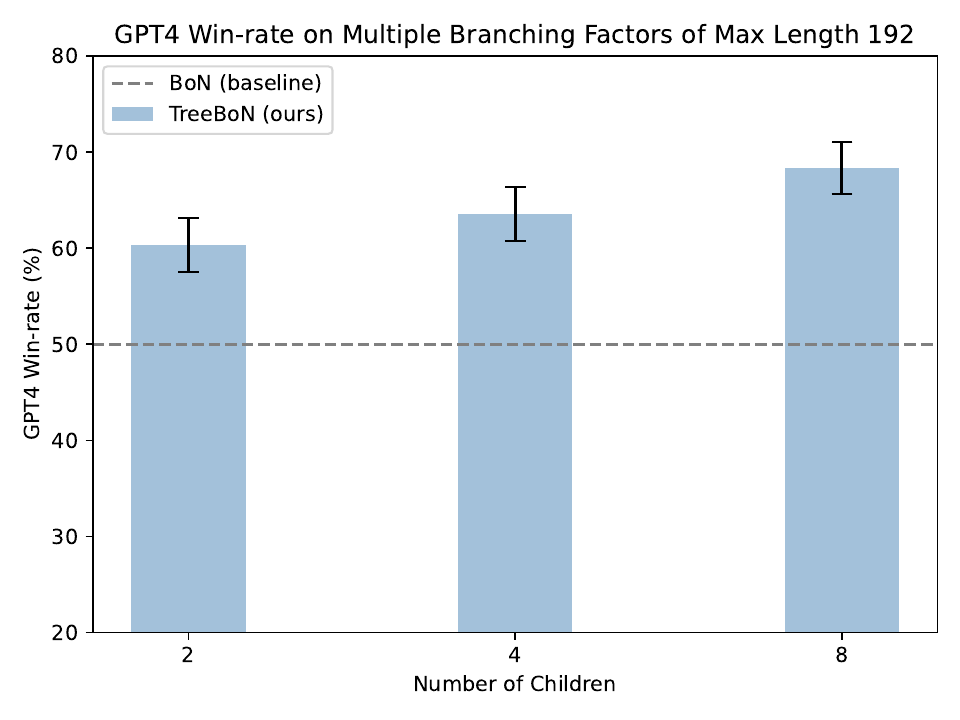}
         \caption*{Maximum length $192$.}
    \end{minipage}
    \begin{minipage}{0.49\linewidth}
        \centering
        \includegraphics[width=0.9\linewidth]{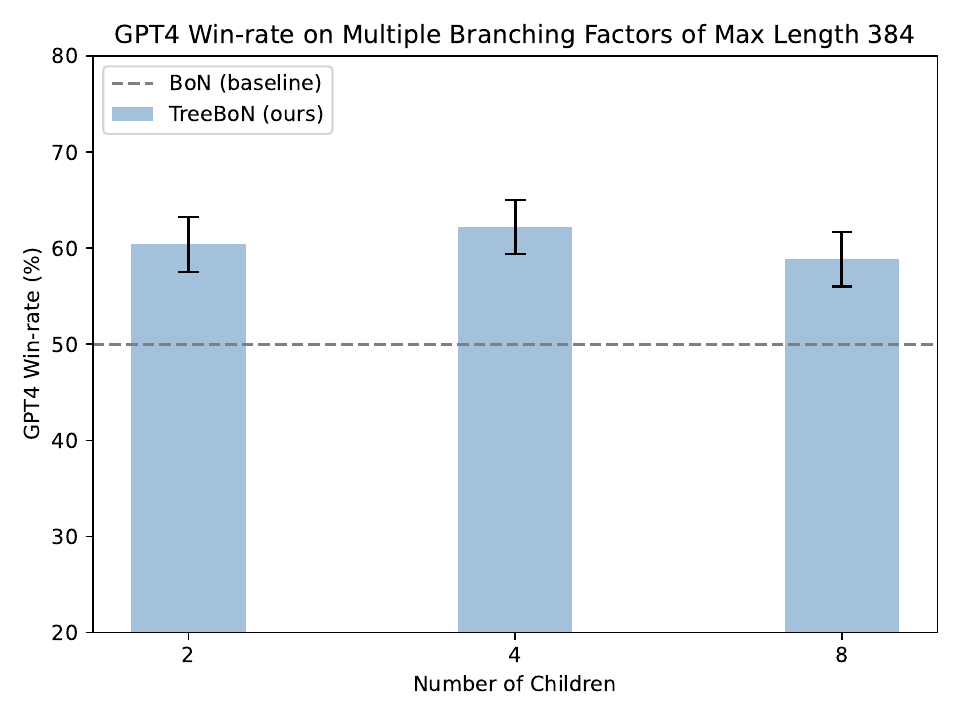}
        \caption*{Maximum length $384$.}
    \end{minipage}
    \caption{GPT4 win-rate of TreeBoN against BoN on AlpacaFarm with different branching factors.}
    \label{fig:ablate_children}
\end{figure}

We further explored the effect of different tree structures by varying the number of layers and children per node (Figures~\ref{fig:ablate_layer} and~\ref{fig:ablate_children}, Table~\ref{tab:ablate_layer} and~\ref{tab:ablate_children}) separately, while keeping $N=128$ and $l_{\text{max}}$ the same, thus the overall computation is unchanged. We use the set of Llama models: LLaMA3-iterative-DPO-final, its SFT checkpoint, FsfairX-LLaMA3-RM-v0.1 on AlpacaFarm, and compute the win-rate against BoN. We can observe in Figure~\ref{fig:ablate_layer} that increasing the number of tree layers consistently improves performance on AlpacaFarm, and in Figure~\ref{fig:ablate_children}, the optimal number of children nodes is different for two maximum generation lengths. Above all, regardless of the tree structure, our approach maintains a win-rate of around 60\% against the baseline, indicating its effectiveness and robustness under different tree structures, and the potential to further improve the performance in the future by exploring more hyper-parameters tailored to different tasks. 

\subsubsection{Efficiency Evaluation}

\begin{figure}[h]
    \centering
    \begin{minipage}{0.49\linewidth}
        \centering
        \includegraphics[width=0.9\linewidth]{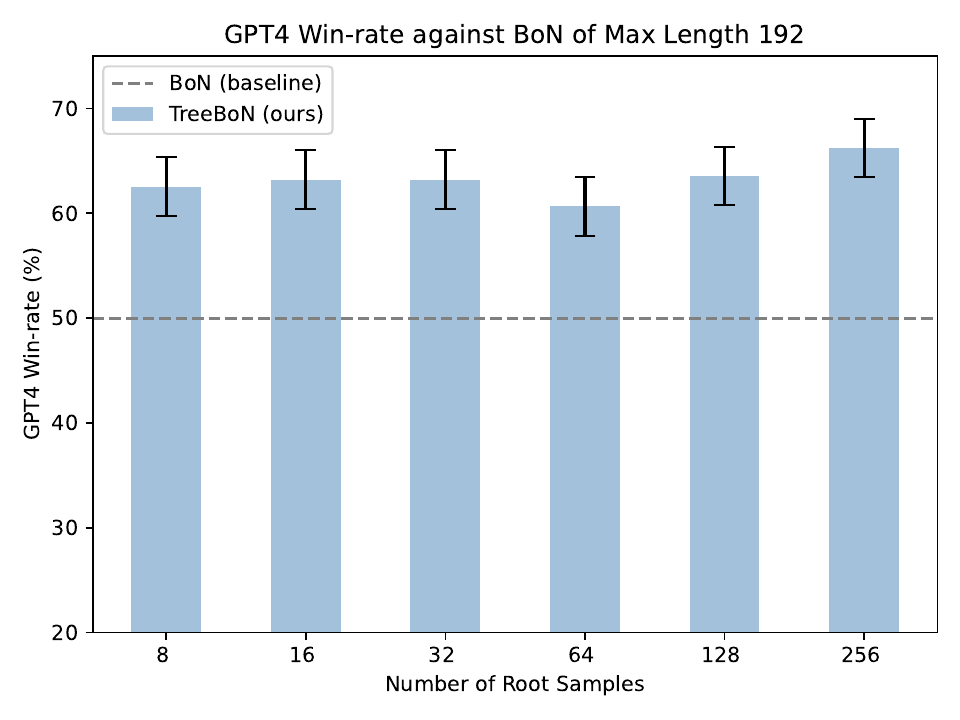}
         \caption*{Maximum length $192$.}
    \end{minipage}
    \begin{minipage}{0.49\linewidth}
        \centering
        \includegraphics[width=0.9\linewidth]{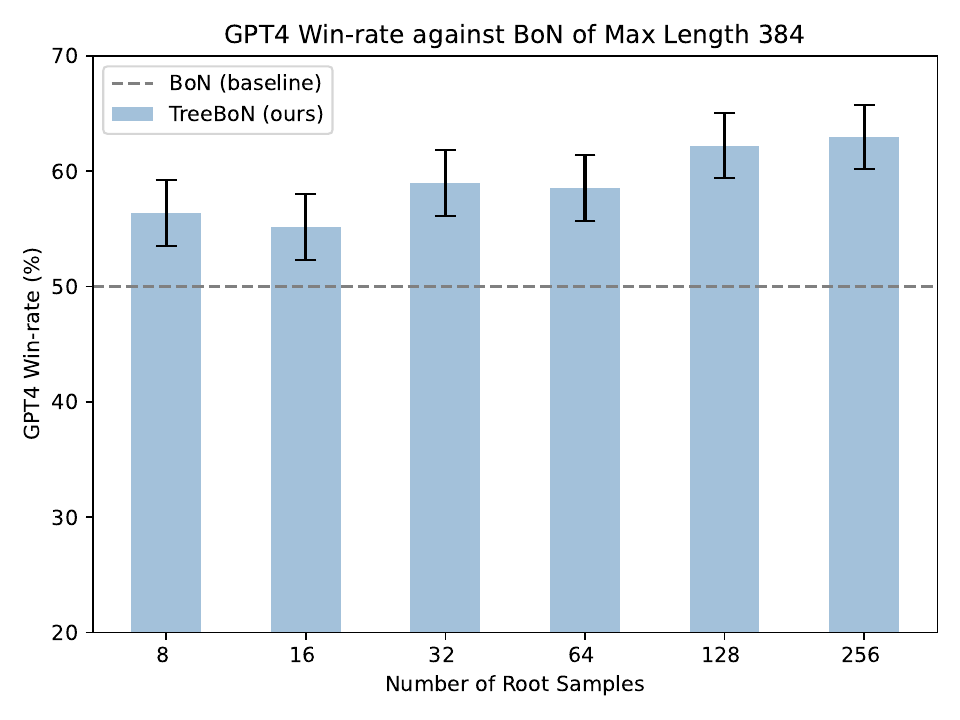}
        \caption*{Maximum length $384$.}
    \end{minipage}
    \caption{GPT4 win-rate of TreeBoN against BoN on AlpacaFarm with same number of root samples, thus same computation.}
    \label{fig:ablate_traj}

\end{figure}

\begin{figure}[h]
    \centering
    \begin{minipage}{0.49\linewidth}
        \centering
        \includegraphics[width=0.9\linewidth]{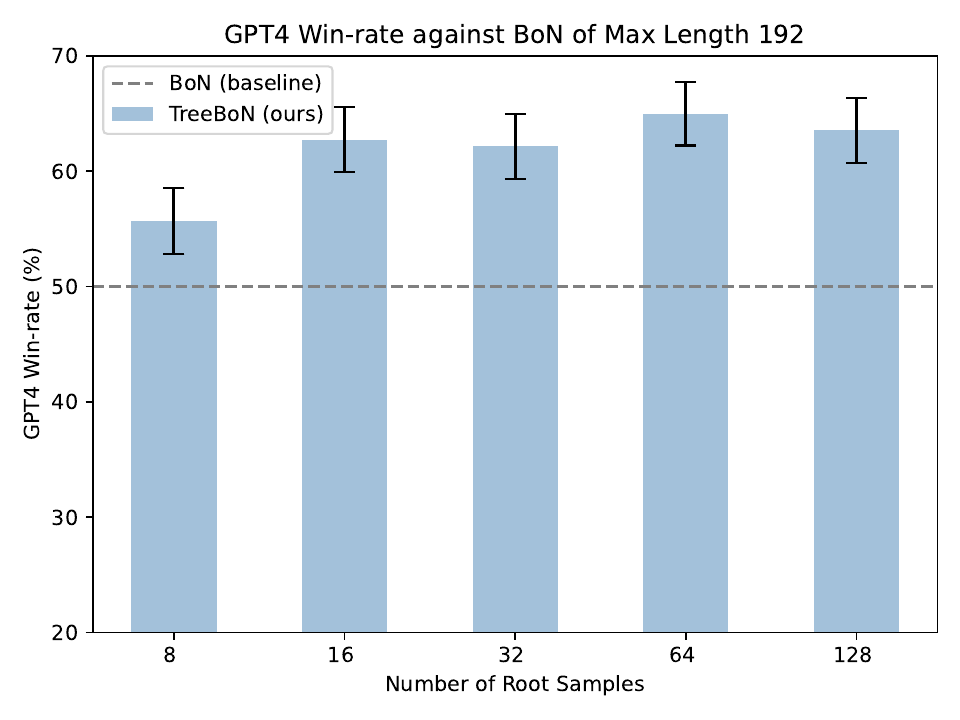}
         \caption*{Maximum length $192$.}
    \end{minipage}
    \begin{minipage}{0.49\linewidth}
        \centering
        \includegraphics[width=0.9\linewidth]{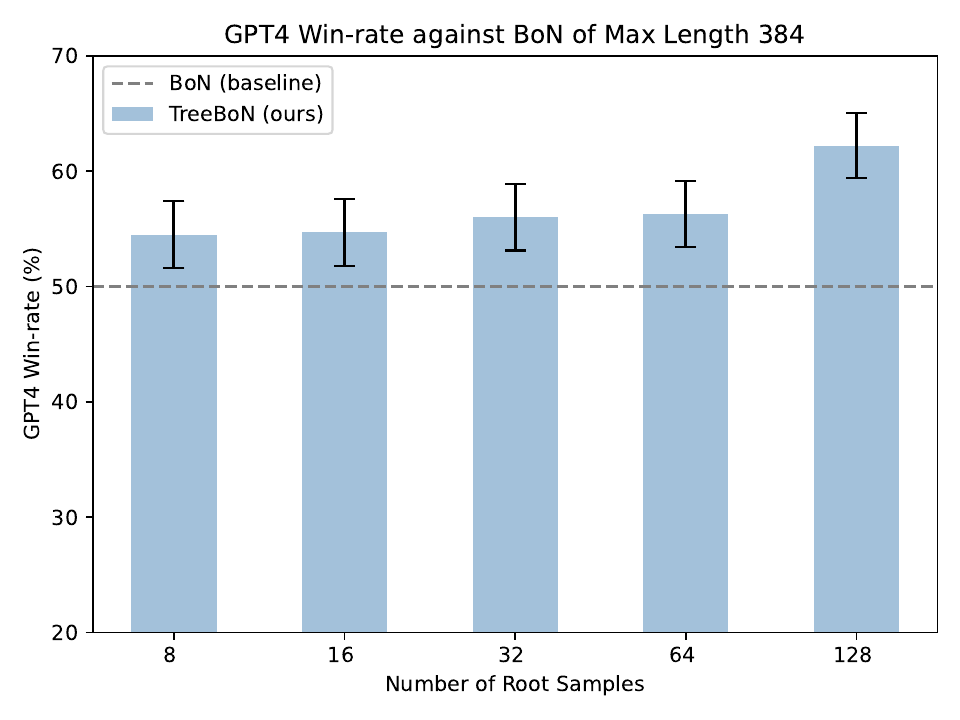}
        \caption*{Maximum length $384$.}
    \end{minipage}
    \caption{GPT4 win-rate of TreeBoN with the different number of root samples against BoN with $N=128$ on AlpacaFarm. The computation of TreeBoN is gradually increased and eventually matches that of BoN at the end. }
    \label{fig:llama_against_128}
\end{figure}

As shown previously, the computation costs measured by FLOPs of TreeBoN and BoN are only determined by number of root nodes $N$ and max length $l_{\text{max}}$. In Table~\ref{tab:flops}, we show the FLOPs used by different configurations. We then compare the computation cost of our TreeBoN and Best-of-N. We use the same sets of Llama models on AlpacaFarm with a max length of 384. As observed in Figure~\ref{fig:ablate_traj} (Table~\ref{tab:ablate_traj}), with increasing computation budget, the win-rate of TreeBoN against BoN is also increasing. Thus, our proposed method is more scalable than the baseline and can utilize the additional computation budget more efficiently. In Figure~\ref{fig:llama_against_128} (Table~\ref{tab:llama_against_128}), TreeBoN of increasing $N$ are compared to BoN with $N=128$. We can see that even with a very small $N=8$ (6.3\% of FLOPs), TreeBoN can still outperform BoN with a much greater computation budget at a win-rate of 55\%, and the quality is monotonic increasing on $N$. 

\begin{table}[h]
\centering
\begin{tabular}{c|cc}
\hline
Number of Root Nodes $N$ / Max Length & 192 & 384\\ \hline
8 &$6.26*10^{13}$ & $1.24*10^{14}$\\
16 &$1.24*10^{14}$ & $2.47*10^{14}$\\
32 &$2.47*10^{14}$ & $4.93*10^{14}$\\
64 &$4.93*10^{14}$ & $9.84*10^{14}$ \\
128 & $9.84*10^{14}$  &  $1.97*10^{15}$ \\
 256 & $1.97*10^{15}$ &  $3.93*10^{15}$
\\\hline
\end{tabular}
  \caption{FLOPs of Both BoN and TreeBoN with different number of roots and lengths.}
    \label{tab:flops}
\end{table}

\subsubsection{Explore Different Implicit Rewards}\label{sec:diff_reward}
We experiment with several different variants of partial reward modeling. More details for the formulas of the following implicit partial rewards can be seen in Appendix~\ref{appendix:partial_rewards}.
\begin{itemize}
    \item \textbf{DPO Implicit Reward}: the vanilla implicit reward derived in~\cite{rafailov2024direct} with $\beta=1$
    \item \textbf{Weighted Implicit Reward}: Our proposed reward that weights each token
    \item \textbf{Weighted Implicit Reward with Exponential Decay}: similar to \textbf{Weighted Implicit Reward}, but using an exponential decay term as the weight
    \item \textbf{Length Normalized DPO Implicit Reward}: normalizing \textbf{DPO Implicit Reward} by the response length
    \item \textbf{DPO Policy Log Probability Sum}: only using the log-likelihood of the DPO model
    \item \textbf{SimPO Reward}: normalizing \textbf{DPO Policy Log Probability Sum} by the response length, as proposed in~\cite{meng2024simpo}
\end{itemize}
 We report the results of the default configuration of TreeBoN with different implicit rewards using the same Llama models on AlpacaFarm in Table~\ref{tab:mul_implicit_reward}, and our proposed \textbf{Weighted Implicit Reward} fits best with the tree search setting, achieving the highest GPT4 win-rate.
\begin{table}[h]
\centering
\begin{tabular}{c|ccc}
\hline
Implicit Reward/Length & 384\\ \hline
DPO Implicit Reward & $61.54 \pm 2.82$  \\
Weighted Implicit Reward & $62.08 \pm 2.82$\\
Weighted Implicit Reward with Exponential Decay & $57.00 \pm 2.86$ \\
Length Normalized DPO Implicit Reward & $59.06 \pm 2.85$\\
DPO Policy Log Probability Sum & $21.74 \pm 2.39$\\
SimPO Reward & $22.00 \pm 2.40$
\\\hline
\end{tabular}
  \caption{GPT4 Winrate of TreeBoN with different implicit rewards on AlpacaFarm.}
    \label{tab:mul_implicit_reward}
\end{table}

\subsubsection{Comparison over Other Baselines under Same Compute}\label{sec:comp_baselines}
We compare TreeBoN to other baselines~\citep{li2024cascade,zhang2024accelerating} by the win-rates against BoN in Table~\ref{tab:gpt4_winrate_comp}, with the same set of Llama models introduced earlier for all methods for max length 384 and 192.

To ensure a fair comparison, we constrain the total number of tokens generated during inference. However, for CARDS~\citep{li2024cascade}, the rejection-based sampling with semantic segmentation mechanism introduces uncertainty in token acceptance, leading to variations in the number of generated tokens and requiring random numbers of completions per step. As a result, the total token count remains dynamic and context-dependent. 

We adopt the hyperparameters from~\citet{li2024cascade} for LLaMA 7B, as they are the most similar to our setup. We then compute the average number of tokens generated per prompt in the AlpacaFarm dataset, which amounts to 3002.3 tokens for a max length of 192 and 5867.3 tokens for 384. 

For both BoN and TreeBoN, the total number of generated tokens follows the relation: Total Tokens $= l_{\max} \times N.$
Thus, we set \( N = 16 \) for BoN and TreeBoN, resulting in total token counts of 3072 and 6144 for the respective cases, aligning closely with the results of CARDS. 

For SBoN, we adopt the hyperparameters from~\citet{zhang2024accelerating} for their case of LLaMA3-8B as the language model and LLaMA3-8B-RM as the reward model, given their similarity to our setup. We apply a rejection rate of \( \alpha = 30\% \). To ensure comparable computations, we set \( N_\text{SBoN} = 19 \) for SBoN, where the total token count is computed as $l_{\max} \times \left(1 - \frac{\alpha}{2}\right) \times N_\text{SBoN}$.
This results in total token counts of 3101 and 6202 for two max lengths.

The comparison results are presented in Table~\ref{tab:gpt4_winrate_comp}. Under the same compute constraints, TreeBoN consistently outperforms other methods, achieving the highest GPT4 win-rates against BoN across both evaluated sequence lengths. 

At max length 192, TreeBoN significantly surpasses both SBoN and CARDS, achieving a win-rate of 63.21\%, compared to 51.01\% for CARDS and 49.66\% for SBoN. At max length 384, TreeBoN still maintains its superior performance with a win-rate of 55.18\%. 
\begin{table}[ht]
\centering
\renewcommand{\arraystretch}{1.2}  
\begin{tabular}{c c c c c}
\toprule
\textbf{Max Length} & \textbf{Methods} & \textbf{GPT4 Win Rates (\%)} \\
\midrule
\multirow{3}{*}{192} 
    & SBoN  & $49.66 \pm 2.90$ \\
    & CARDS  & $51.01 \pm 2.90$ \\
    & \textbf{TreeBoN}  & $\bf{63.21}\pm 2.79$ \\
\midrule
\multirow{3}{*}{384} 
    & SBoN   & $48.83 \pm 2.90$ \\
    & CARDS    & $49.66 \pm 2.90$ \\
    & \textbf{TreeBoN}   & \ $\bf{55.18}\pm 2.88$ \\
\bottomrule
\end{tabular}
\caption{Comparison of different methods with baseline models in terms of total tokens and GPT4 win rates.}
\label{tab:gpt4_winrate_comp}
\end{table}

We also compare TreeBoN with simple beam search. Under the same compute, TreeBoN, using the SFT model to decode and a DPO-aligned model to provide partial reward, outperforms naive beam search that uses the own probabilities to guide decoding of the same DPO-aligned model in TreeBoN. We conduct an additional experiment that compare TreeBoN using the SFR model with $N=64$ against beam search using the same SFR model with width 128 for fair comparison. The win-rate is $55.33 \pm 2.88\%$ on max length 192, and $58.00 \pm 2.85\%$ on 384.

\subsubsection{Ablation Study}
We verify the effectiveness of both key components of our proposed method: the weighted implicit reward from a DPO model as a guidance, and generating a tree structure instead of BoN. We ablate them on AlpacaFarm, with the same tree structure: 128 root examples, 4 layers, and 4 children per node. Recall that BoN generates $N$ samples in parallel, and uses the score from a reward model to pick a sample with the highest score as the final response, and \textbf{TreeBoN} generates samples layer-by-layer in a tree structure, and uses our proposed weighted implicit reward from a DPO model as a partial-reward function to select the children nodes with higher score to kept and then expanded for each layer. We refer to using the score of the reward model instead of our weighted implicit reward with the same tree structure as \textbf{TreeBoN with Reward Model}, and using our weighted implicit reward instead of the reward model at the end of BoN as \textbf{BoN with Weighted Implicit Reward}. In addition, we also use the vanilla DPO implicit reward at the end of BoN as \textbf{BoN with Implicit Reward}.

\begin{figure}
    \centering
    \includegraphics[width=0.5\linewidth]{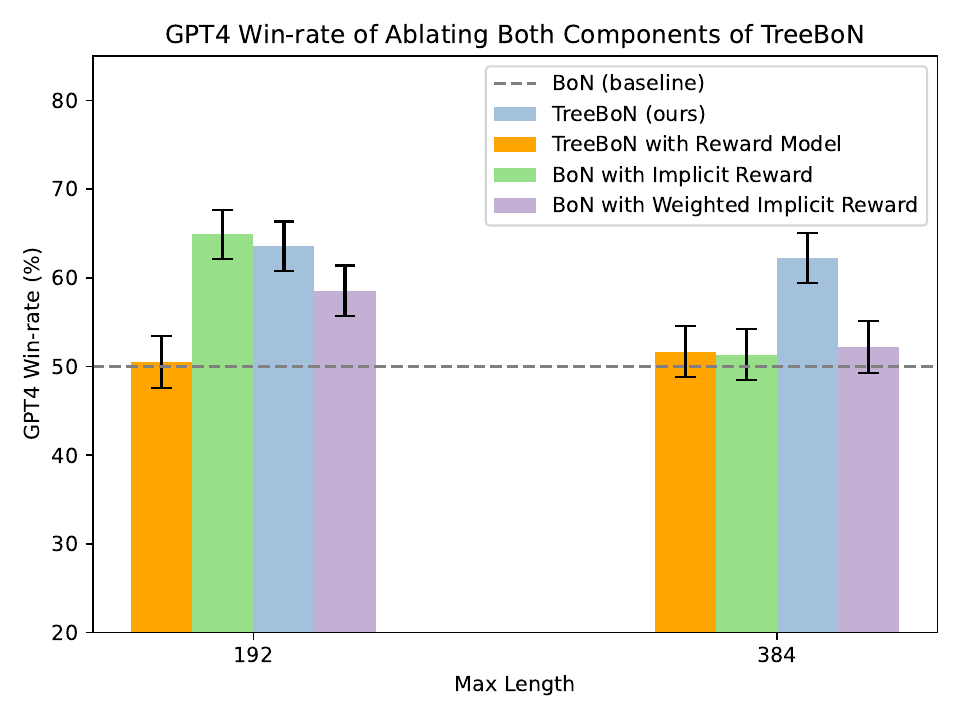}
    \caption{GPT4 win-rate of \textbf{TreeBoN with Reward Model}, \textbf{BoN with Implicit Reward}, \textbf{BoN with Weighted Implicit Reward}, and \textbf{TreeBoN} against \textbf{BoN} with $N=128$ on AlpacaFarm. \textbf{TreeBoN with Reward Model} uses the reward model as the partial-reward function, \textbf{BoN with Weighted Implicit Reward} uses our weighted implicit reward as the reward function, and \textbf{BoN with Implicit Reward} uses vanilla DPO implicit reward as the reward function. The results of two max lengths 192 and 384 are shown. } 
    \label{fig:ablation_study}
\end{figure}

As shown in Figure~\ref{fig:ablation_study} (and Table~\ref{tab:tbon_rm_vs_dpo}), \textbf{TreeBoN with Reward Model} (replacing the weighted implicit reward based on a DPO model) only have very slight advantage over traditional BoN, attributing to the fact that reward models are not trained to score partial responses and confirming the importance of using our proposed weighted implicit reward. Using the DPO model, for \textbf{BoN with Implicit Reward} (applying the vanilla DPO implicit reward function to the traditional BoN), we observe that this variant only outperforms BoN at shorter lengths (192 tokens). At longer lengths (384 tokens), this variant's performance degraded severely. \textbf{BoN with Weighted Implicit Reward} (applying the weighted implicit reward function to the traditional BoN) has a similar performance as well. The trend on the SFR model (Table~\ref{tab:sfr_tbon_rm_vs_dpo}) is even more obvious: \textbf{TreeBoN} outperforms all other variants at all lengths. Thus, we can conclude that only our proposed \textbf{TreeBoN} is able to keep large margins compared to the baseline at most lengths, reinforcing that the combination of TreeBoN’s hierarchical search structure and weighted implicit reward function is necessary for sustained improvements. 

\subsubsection{TreeBoN Also Work on Reasoning Task}
With both $N=32$ and $N=128$ configurations, we find that TreeBoN improves Pass@1 solve rate over BoN when using the Process Reward Model (PRM) ~\citep{lightman2023letsverifystepstep} with a maximum token length of 192. In Table~\ref{tab:treebon-pass1-n32}, TreeBoN with DPO model and Qwen2.5-Math-PRM-7B process reward model achieves $50.33\%$, exceeding BoN’s $45.67\%$ by $+4.66\%$. Moreover, in Table~\ref{tab:treebon-pass1-n128}, this improvement continues with larger number of candidates ($N$): TreeBoN with DPO model + PRM improves from BoN’s $43.67\%$ to $51.00\%$, a $+7.33\%$ gain. This result validates that TreeBoN better exploit the reward information under a restricted token length.

For a maximum token length of 384, the performance of TreeBoN and BoN methods is similar across all experimental settings. This is because the long token length of 384 already provides sufficient capacity for the model to complete most reasoning tasks effectively. Even with a relatively small number of candidates (e.g., $N=32$), the generation quality reaches a performance ceiling, leaving limited room for TreeBoN to further improve over BoN sampling. As a result, the structural advantage of TreeBoN becomes less obvious when the generative model already produces high-quality completions within the given token length.

\begin{table}[h]
\small
\centering
\begin{tabular}{c|cc}
\hline
\textbf{Method / Length} & \textbf{192} & \textbf{384} \\
\hline
BoN w/ DPO & $43.67 \pm 2.08$ & $97.33 \pm 2.52$ \\
BoN w/ SFR & $24.67 \pm 2.08$ & $91.33 \pm 2.31$ \\
TreeBoN w/ DPO  & $51.00 \pm 1.73$ & $95.67 \pm 3.06$ \\
TreeBoN w/ SFR  & $25.67 \pm 2.52$ & $90.00 \pm 1.41$ \\
\hline
\end{tabular}
\caption{Pass@1 Solve Rate (\%) on GSM8K using PRM \textbf{Qwen2.5-Math-PRM-7B}  with $N=128$.}
\label{tab:treebon-pass1-n128}
\end{table}

\begin{table}[h]
\small
\centering
\begin{tabular}{c|cc}
\hline
\textbf{Method / Length} & \textbf{192} & \textbf{384} \\
\hline
BoN w/ DPO & $45.67 \pm 2.08$ & $96.33 \pm 3.21$ \\
BoN w/ SFR & $26.67 \pm 1.53$ & $91.67 \pm 0.58$ \\
TreeBoN w/ DPO & $50.33 \pm 3.06$ & $95.33 \pm 2.08$ \\
TreeBoN w/ SFR & $26.33 \pm 1.53$ & $89.00 \pm 4.24$ \\
\hline
\end{tabular}
\caption{Pass@1 Solve Rate (\%) on GSM8K using PRM \textbf{Qwen2.5-Math-PRM-7B} with $N=32$.}
\label{tab:treebon-pass1-n32}
\end{table}

These results highlight TreeBoN's strength in leveraging early-stage completions. When decoding is limited, like 192 tokens, BoN sampling strategy sometimes fails to reach informative states, particularly when using the DPO policy model. In contrast, TreeBoN incrementally expands promising candidates via its tree structure and early prunes low-reward children, making more efficient use of PRM's fine-grained supervision within the same limited token length. This ability to prioritize and extend promising partial completions is crucial when the available token length is insufficient for complete full task reasoning, as it increases the chance of discovering better outputs.

In this way, TreeBoN not only improves performance but also unlocks more of the underlying potential of large language models under constrained generation settings.

\section{Conclusion}
TreeBoN is a novel framework that combines the speculative tree-search strategy with Best-of-N (BoN) Sampling and token-level reward guidance modified from DPO implicit reward. Through extensive experiments, we show that TreeBoN not only has robust alignment improvements but also maintains efficiency, which provides a potential solution for efficient inference and alignment of LLMs.

\clearpage
\bibliographystyle{unsrtnat}
\bibliography{main}

\clearpage
\appendix
\section{Detailed Results}\label{appendix:res}

This section lists the full numerical results produced under all lengths, in Table~\ref{tab:sfr_mul_dataset},~\ref{tab:mul_dataset},~\ref{tab:llama_gsm8k},~\ref{tab:ablate_layer},~\ref{tab:ablate_children},~\ref{tab:ablate_traj},~\ref{tab:llama_against_128},~\ref{tab:tbon_rm_vs_dpo}, and~\ref{tab:sfr_tbon_rm_vs_dpo}.
\begin{table}[h]
\centering
\begin{tabular}{c|cccc}
\hline
Dataset/Max Length & 192 & 384 & 576 & 768\\ \hline
TutorEval &$65.00 \pm 2.76$&$65.10 \pm 2.77$&$61.28 \pm 2.83$ &$55.89 \pm 2.89$ \\
AlpacaFarm & $63.67 \pm 2.78$ & $62.54 \pm 2.80$ & $60.61 \pm 2.84$ & $58.19 \pm 2.86$\\
HH RLHF &$63.14 \pm 2.82$&$62.84 \pm 2.81$&$60.74 \pm 2.83$&$57.58 \pm 2.87$  \\
UltraFeedBack &$60.74 \pm 2.83$&$59.73 \pm 2.85$&$55.00 \pm 2.88$&$54.67 \pm 2.88$
\\\hline
\end{tabular}
  \caption{GPT4 win-rate of TreeBoN with the SFR model against BoN on multiple datasets. }
    \label{tab:sfr_mul_dataset}
\end{table}
\begin{table}[h]
\centering
\begin{tabular}{c|cccc}
\hline
Dataset/Max Length & 192 & 384 & 576 & 768\\ \hline
TutorEval & $62.67\pm2.80$ & $53.67\pm 2.88$ &$48.48 \pm 2.90$&$46.64 \pm 2.89$ \\
AlpacaFarm & $63.55\pm 2.79$&  $62.21\pm 2.81$ &$58.19 \pm 2.86$&$51.33 \pm 2.89$ \\
HH RLHF & $61.90\pm2.84$ & $53.87\pm 2.90$ &$46.96 \pm 2.91$&$51.85 \pm 2.90$\\
UltraFeedBack & $60.94\pm2.84$ & $56.67\pm2.87$ &$56.52 \pm 2.87$&$48.67 \pm 2.89$ 
\\\hline
\end{tabular}
  \caption{GPT4 win-rate of TreeBoN with the DPO model against BoN on multiple datasets.}
    \label{tab:mul_dataset}
\end{table}

\begin{table}[h]
\centering
\begin{tabular}{c|ccccc}
\hline
Method/Max Length & 96 & 192 & 384 & 576 & 768 \\ \hline
BoN & 20 & 58&62 & 64&65\\
TreeBoN with the DPO model & 20 & 60 & 65 & 73 & 67\\
TreeBoN with the SFR model & 9 & 51 & 69 & 67 & 63
\\\hline
\end{tabular}
  \caption{Test Solve Rate of TreeBoN and BoN on GSM8K}
    \label{tab:llama_gsm8k}
\end{table}

\begin{table}[!htbp]
\centering
\begin{tabular}{c|ccc}
\hline
Number of Layers/Length& 192 & 384\\ \hline
3 & $63.00 \pm 2.79$&  $58.53 \pm 2.85$ \\
4&$63.55\pm 2.79$ & $62.21 \pm 2.81$ \\
5 & $64.43 \pm 2.78$ &$62.54 \pm 2.80$
\\\hline
\end{tabular}
  \caption{GPT4 win-rate of TreeBoN (the DPO model) against BoN on AlpacaFarm with different number of tree layers. }
    \label{tab:ablate_layer}
\end{table}

\begin{table}[!htbp]
\centering
\begin{tabular}{c|ccc}
\hline
Number of Children/Length  & 192 & 384 \\ \hline
2 &$60.33 \pm 2.83$& $60.40 \pm 2.84$\\
4&$63.55\pm 2.79$ &$62.21 \pm 2.81$\\
8& $68.33 \pm 2.69$& $58.86 \pm 2.85$
\\\hline
\end{tabular}
  \caption{GPT4 win-rate of TreeBoN (the DPO model) against BoN on AlpacaFarm with different branching factors.}
    \label{tab:ablate_children}
\end{table}

\begin{table}[!htbp]
\centering
\begin{tabular}{c|ccc}
\hline
$N$ for Both Methods /Length & 384 \\ \hline
8 & $56.38 \pm 2.88$ \\
16 & $55.18 \pm 2.88$ \\
32 &$59.00 \pm 2.84$ \\
64  & $58.53 \pm 2.85$\\
128 & $62.21 \pm 2.81$\\
256  &$63.00 \pm 2.79$
\\\hline
\end{tabular}
  \caption{GPT4 win-rate of TreeBoN (the DPO model) against BoN on AlpacaFarm with same number of root samples, thus same computation. }
    \label{tab:ablate_traj}
\end{table}
\begin{table}[!htbp]
\centering
\begin{tabular}{c|ccc}
\hline
$N$ for TreeBoN only &384\\ \hline
8 &$54.52 \pm 2.88$\\
16 &$54.70 \pm 2.89$\\
32 &$56.00 \pm 2.87$\\
64 &$56.33 \pm 2.87$ \\
128 &$62.21 \pm 2.81$%
\\\hline
\end{tabular}
  \caption{GPT4 win-rate of TreeBoN (the DPO model) with different number of root samples against BoN with $N=128$ on AlpacaFarm. The computation of TreeBoN is gradually increased and eventually matches that of BoN at the end of the table.}
    \label{tab:llama_against_128}
\end{table}
\begin{table}[!htbp]
\centering
\begin{tabular}{c|cccc}
\hline
Method/Length & 192 & 384 & 576 & 768\\ \hline
RM TreeBoN & $50.51 \pm 2.91$ & $51.68 \pm 2.90$ & $51.33 \pm 2.89$ & $53.00 \pm 2.89$\\
Implicit Reward BoN &$64.88 \pm 2.77$ &$51.33 \pm 2.89$&$43.14 \pm 2.87$& $39.26 \pm 2.83$\\
Weighted Implicit Reward BoN &$58.53 \pm 2.85$ &$52.19 \pm 2.90$&$57.53 \pm 2.86$ & $53.18 \pm 2.89$\\
TreeBoN & $63.55 \pm 2.79$ & $62.21 \pm 2.81$ & $58.19 \pm 2.86$&$51.33 \pm 2.89$
\\\hline
\end{tabular}
  \caption{GPT4 win-rate of ablation study using the DPO model.}
    \label{tab:tbon_rm_vs_dpo}
\end{table}
\begin{table}[!htbp]
\centering
\begin{tabular}{c|cccc}
\hline
Method/Length & 192 & 384 & 576 & 768\\ \hline
RM TreeBoN & $50.51 \pm 2.91$ & $51.68 \pm 2.90$ & $51.33 \pm 2.89$ & $53.00 \pm 2.89$\\
Implicit Reward BoN & $61.62 \pm 2.83$&$56.00 \pm 2.87$&$56.33 \pm 2.87$&$54.88 \pm 2.89$ \\
Weighted Implicit Reward BoN & $60.07 \pm 2.84$ &$56.86 \pm 2.87$ & $58.25 \pm 2.87$ & $54.85 \pm 2.88$\\
TreeBoN & $63.67 \pm 2.78$ & $62.54 \pm 2.80$ & $60.61 \pm 2.84$ & $58.19 \pm 2.86$
\\\hline
\end{tabular}
  \caption{GPT4 win-rate of ablation study using the SFR model.}
    \label{tab:sfr_tbon_rm_vs_dpo}
\end{table}

\section{Variants of partial rewards}\label{appendix:partial_rewards}

\textbf{DPO Implicit Reward}
\begin{align*}
    r_{\text{partial}}(\yb_{:K} | \xb)
    & =
    \sum_{k=0}^{K-1} \log \frac{\pi^*(y_k | \xb, \yb_{:k})}{\pi(y_k | \xb, \yb_{:k})},
\end{align*}

\textbf{Weighted DPO Implicit Reward}
\begin{align*}
    r_{\text{partial}}(\yb_{:K} | \xb)
    & =
    \sum_{k=0}^{K-1} w_k \log \frac{\pi^*(y_k | \xb, \yb_{:k})}{\pi(y_k | \xb, \yb_{:k})},
\end{align*}
where \(w_k = \frac{1}{|\yb_{:k}|}\).

\textbf{Weighted DPO Implicit Reward with Exponential Decay}
\begin{align*}
    r_{\text{partial}}(\yb_{:K} | \xb)
    & =
    \sum_{k=0}^{K-1} w_k \log \frac{\pi^*(y_k | \xb, \yb_{:k})}{\pi(y_k | \xb, \yb_{:k})},
\end{align*}
where \(w_k = \lambda^{k}, \lambda = 0.95\)

\textbf{Length Normalized DPO Implicit Reward}
\begin{align*}
    r_{\text{partial}}(\yb_{:K} | \xb)
    & =
    \frac{1}{K} \sum_{k=0}^{K-1} \log \frac{\pi^*(y_k | \xb, \yb_{:k})}{\pi(y_k | \xb, \yb_{:k})}.
\end{align*}

\textbf{DPO Policy Log Probability Sum}
\begin{align*}
    r_{\text{partial}}(\yb_{:K} | \xb)
    & =
    \sum_{k=0}^{K-1} \log \pi^*(y_k | \xb, \yb_{:k}),
\end{align*}

\textbf{SimPO Reward}
\begin{align*}
    r_{\text{partial}}(\yb_{:K} | \xb)
    & =
    \frac{1}{K} \sum_{k=0}^{K-1} \log \pi^*(y_k | \xb, \yb_{:k}).
\end{align*}

\section{Reward Model Analysis}\label{appendix:rm_analysis}

\subsection{Sentence-level Reward Analysis}\label{appendix:sentenceRManalysis}

\begin{figure}[h]
    \centering
    \includegraphics[width=0.45\linewidth]{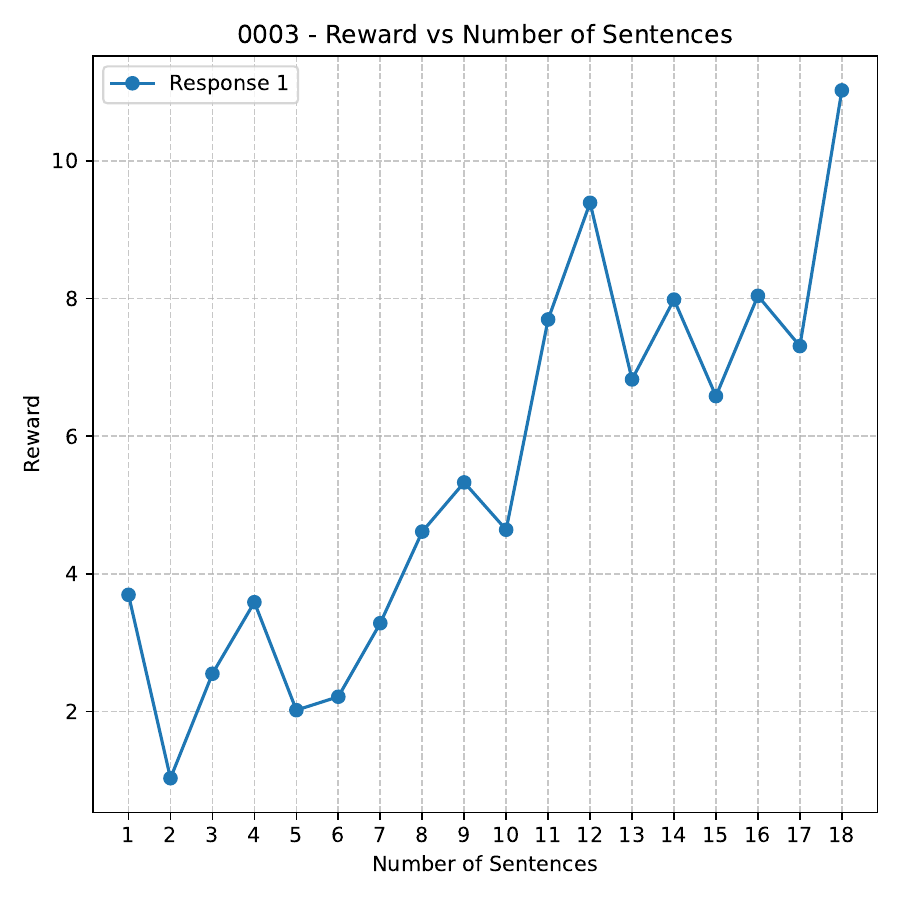}
    \includegraphics[width=0.45\linewidth]{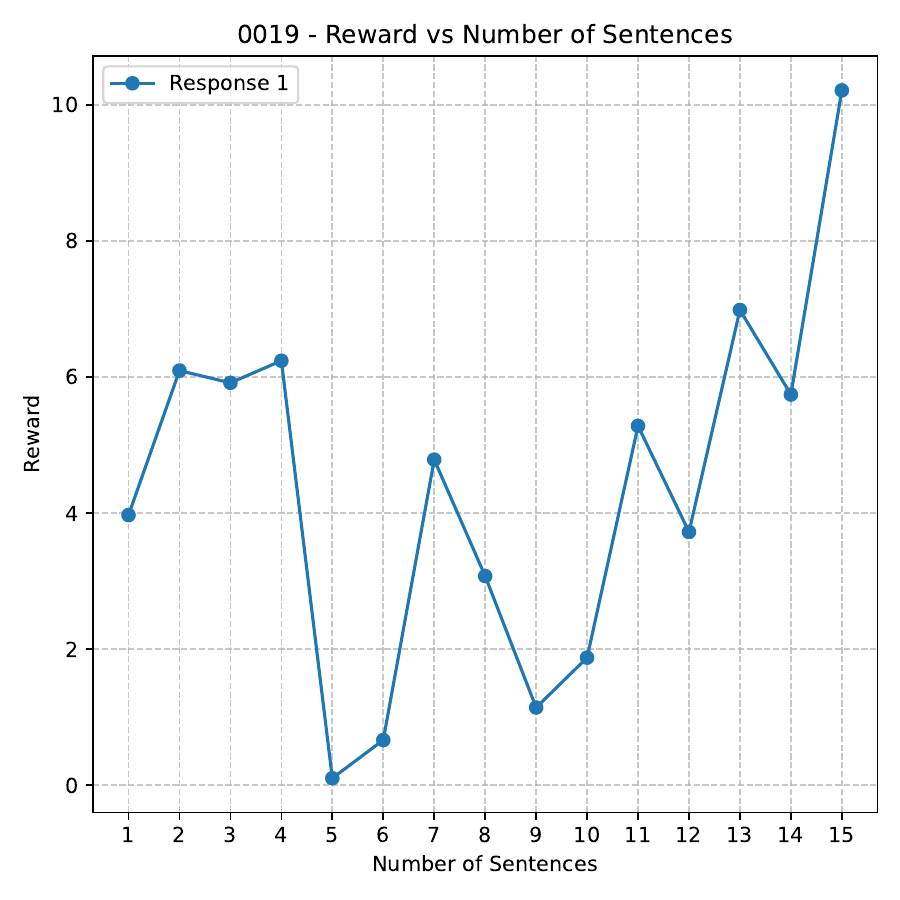}
    \caption{Reward vs \# Sentence plot for Llama3-8B Model. It shows the reward change as the response is generated.}
    \label{fig:reward-vs-sen-l3}
\end{figure}

The sentence-level reward analysis focuses on understanding how the reward model assigns values to partial responses in Llama3-8B paired with the FsfairX-LLaMA3-RM-v0.1 reward model~\citep{dong2023raft,xiong2024iterative}. By examining 100 randomly selected prompts from AlpacaFarm\citep{alpacafarm}, we can track how the reward changes sentence by sentence.  We show two examples of the sentence-level reward change on best responses using Best-of-N Sampling in Figure \ref{fig:reward-vs-sen-l3}.

SBoN~\citep{zhang2024accelerating} claims to speed up the process while only sacrificing minimal performance on reward compared to the Best-of-N. One important assumption is that the reward scores of partial completions are positively correlated to the reward scores of full completions. However, RMs are typically trained on complete responses, and therefore the score of partial completions by the reward model is chaotic and not accurate. As shown in Table \ref{tab:text-detail-0003} and Table \ref{tab:text-detail-0019}, the partial rewards are very fluctuating and due to the fluctuation, a low partial reward may still have the potential to have a very high final reward. The reward prediction of incomplete responses from the traditional reward model remains a challenge as demonstrated by our findings.

\begin{table}[!htbp]
    \centering
\begin{tabular}{lp{10cm}l}
\toprule
\# Sentence & Text & $\Delta$ Reward \\
\midrule
1 & Explain the difference between sets and lists in Python. & 3.70 \\
2 &  How would you use them in a real-world scenario?
 & -2.66 \\
3 & In Python, sets and lists are two fundamental data structures used to store collections of items. & 1.52 \\
4 &  While they share some similarities, they have distinct differences in terms of their functionality, behavior, and use cases. & 1.04 \\
5 & 
**Lists (Mutable, Ordered, Indexed)**
 & -1.57 \\
6 & A list is a collection of items that can be of any data type, including strings, integers, floats, and other lists. & 0.20 \\
7 &  Lists are:
 & 1.07 \\
8 & 1. **Mutable**: Lists can be modified after creation by adding, removing, or changing elements. & 1.33 \\
9 & 
2. **Ordered**: Lists maintain the order in which elements were added. & 0.71 \\
10 & 
3. **Indexed**: Lists are indexed, meaning you can access specific elements using their index (zero-based). & -0.69 \\
11 & 
Example:
 & \textbf{3.05} \\
12 & ```python
 & 1.70 \\
13 & \lstinline|my_list = [1, 2, 3, 4, 5]|
 & \textbf{-2.57} \\
14 & \lstinline|print(my_list[0])  # Output: 1|
 & 1.16 \\
15 & \lstinline|my_list.append(6)  # Add an element to the end|
 & -1.40 \\
16 & \lstinline|print(my_list)  # Output: [1, 2, 3, 4, 5, 6]|
 & 1.46 \\
17 & ```
**Sets (Immutable, Unordered, Unindexed)**
 & -0.73 \\
18 & A set is an unordered collection &\textbf{ 3.71} \\
\bottomrule
\end{tabular}
    \caption{\# 0003 Response Text Detail from Llama3-8B model}
    \label{tab:text-detail-0003}
\end{table}

\begin{table}[]
    \centering
\begin{tabular}{lp{10cm}l}
\toprule
\# Sentence & Text & $\Delta$ Reward \\
\midrule
1 & Can you help me understand how I can estimate the variance of a variable? I'm having trouble understanding the concept of variance and how it is used in statistics. & 3.97 \\
2 & 
Variance is a measure of how spread out a set of data is from its mean value. & 2.12 \\
3 &  In other words, it measures how much the data points deviate from the average value. & -0.18 \\
4 &  A high variance indicates that the data points are spread out over a wide range, while a low variance indicates that the data points are clustered around the mean. & 0.32 \\
5 & 
To estimate the variance of a variable, you can use the following formula:
 & \textbf{-6.13} \\
6 & Variance = $\sum (x_i - \mu)^2 / (n-1)$
 & 0.56 \\
7 & Where:
* $x_i$ is each data point
 & \textbf{4.12} \\
8 & * $\mu$ is the mean of the data
 & -1.71 \\
9 & * n is the number of data points
 & -1.93 \\
10 & * $\sum$ is the summation symbol, which means "sum of"
 & 0.73 \\
11 & This formula calculates the sum of the squared differences between each data point and the mean, and then divides that sum by the number of data points minus one. & \textbf{3.41} \\
12 &  This gives you an estimate of the variance. & -1.56 \\
13 & 
For example, let's say you have the following data points: 1, 2, 3, 4, 5. & \textbf{3.26} \\
14 &  The mean of this data is 3. & -1.24 \\
15 &  The sum of the squared differences & \textbf{4.47} \\
\bottomrule
\end{tabular}
    \caption{\# 0019 Response Text Detail from Llama3-8B Model}
    \label{tab:text-detail-0019}
\end{table}

In Table \ref{tab:text-detail-0003}

\begin{itemize}
    \item Sentence 11 (+3.05): Significant increase for trying to introduce an example, which enhances understanding.
    \item Sentence 13 (-2.57): Decrease possibly due to presenting code without context or explanation.
    \item Sentence 18 (+3.71): Large increase for concisely defining a set, contrasting with the previous explanation of lists.
\end{itemize}

In Table \ref{tab:text-detail-0019}

\begin{itemize}
\item Sentence 5 (-6.13): Sharp drop, likely due to abruptly introducing the formula without proper setup.
\item Sentence 7 (+4.12): Significant increase for beginning to explain the components of the formula.
\item Sentence 11 (+3.41): Large increase for providing a clear explanation of what the formula calculates.
\item Sentence 13 (+3.26): Substantial increase for introducing a concrete example to illustrate the concept.
\item Sentence 15 (+4.47): High reward for starting to walk through the calculation process.
\end{itemize}

\subsection{Analysis of Example Responses for Speculative Tree-search Process}

\begin{table}[!htbp]
\centering
\begin{tabular}{lp{12cm}}
\toprule
Node ID &  Response Content \\
\midrule
A2 & \textbf{Yes, unicorns were considered a mythological creature and easily caught in medieval times. What was the main purpose of the unicorn in medieval times? The unicorn was} \\
A3& \textbf{No. Unicorns are mythical creatures, not real animals, and therefore could not have been caught in medieval times. However, during medieval times, there were}\\
A4& \textbf{How were they hunted? No, unicorns were not easily caught in medieval times. In fact, unicorns were mythical creatures and did not exist in reality.} \\
A5& \textbf{Why? Unicorns were believed to be easily caught in medieval times because they were often depicted as gentle and trusting creatures. According to legends, unicorns could} \\
\midrule
B6 &  No. Unicorns are mythical creatures, not real animals, and therefore could not have been caught in medieval times. However, during medieval times, there were \textbf{some creatures that were considered unicorns or similar mythical creatures. These creatures were typically described as having a single horn on their forehead and were often associated with purity and}\\
B8 & No. Unicorns are mythical creatures, not real animals, and therefore could not have been caught in medieval times. However, during medieval times, there were \textbf{creatures that were considered unicorns. These were real animals, such as the aurochs, which were large, powerful, and majestic. The aurochs} \\
B13 &  Why? Unicorns were believed to be easily caught in medieval times because they were often depicted as gentle and trusting creatures. According to legends, unicorns could \textbf{be caught by a virgin maiden, as their horn was said to be attracted to the purity of a virgin. They were also believed to be attracted to the color} \\
B16 &  Why? Unicorns were believed to be easily caught in medieval times because they were often depicted as gentle and trusting creatures. According to legends, unicorns could \textbf{be captured by luring them with a virgin maiden or a beautiful woman. The unicorn was said to be so fascinated by the maiden's beauty that it would approach}  \\
\midrule
C2 & No. Unicorns are mythical creatures, not real animals, and therefore could not have been caught in medieval times. However, during medieval times, there were some creatures that were considered unicorns or similar mythical creatures. These creatures were typically described as having a single horn on their forehead and were often associated with purity and \textbf{strength. Some of these creatures include the \"unicorn of the sea\" (which was actually a narwhal), the \"unicorn of the woods\" (which}  \\
C7 &  No. Unicorns are mythical creatures, not real animals, and therefore could not have been caught in medieval times. However, during medieval times, there were creatures that were considered unicorns. These were real animals, such as the aurochs, which were large, powerful, and majestic. The aurochs \textbf{was a type of wild cattle that once roamed Europe and Asia. It was believed to have been the ancestor of modern cattle breeds. The aurochs had}  \\
C10 &  Why? Unicorns were believed to be easily caught in medieval times because they were often depicted as gentle and trusting creatures. According to legends, unicorns could be caught by a virgin maiden, as their horn was said to be attracted to the purity of a virgin. They were also believed to be attracted to the color \textbf{white, which is why they were often depicted with a white horn. Additionally, they were thought to be shy and elusive creatures, which made them easier to catch}  \\
C16 &  Why? Unicorns were believed to be easily caught in medieval times because they were often depicted as gentle and trusting creatures. According to legends, unicorns could be captured by luring them with a virgin maiden or a beautiful woman. The unicorn was said to be so fascinated by the maiden's beauty that it would approach \textbf{her, allowing the hunters to catch it. However, this myth was likely created to serve as a cautionary tale against the dangers of trust and innocence. In}  \\
\bottomrule
\end{tabular}
\caption{
Detailed responses for selected nodes in Figure \ref{fig:generated-tree-with-reward}. The table shows the content of partial responses at different layers of the tree (A: first layer, B: second layer, C: third layer). Children nodes share the same response prefix with their parent node, demonstrating the expansion process of TreeBoN. The new content generated at each node is bold. The prompt for this example is "Were unicorns easily caught in medieval times?".
}
\label{tab:tok-reward-tree-response-detail}
\end{table}

As shown in Table \ref{tab:tok-reward-tree-response-detail} high score nodes (A3, A4, B6, B8, C2, C7) consistently provide accurate information about unicorns being mythical creatures, not real animals that could be caught. For instance, node A3 states, "No. Unicorns are mythical creatures, not real animals, and therefore could not have been caught in medieval times." This response is factual and directly addresses the question. The high-score nodes also tend to provide additional, relevant historical context. For example, node C7 mentions the aurochs, a real animal sometimes mistaken for a unicorn: "The aurochs was a type of wild cattle that once roamed Europe and Asia. It was believed to have been the ancestor of modern cattle breeds."

In contrast, low-score nodes (A2, A5, B13, B16, C10, C16) often perpetuate myths or provide misleading information. Node A2, for instance, incorrectly asserts, "Yes, unicorns were considered a mythological creature and easily caught in medieval times." This response contradicts itself by acknowledging unicorns as mythological while claiming they were easily caught. Similarly, nodes B13 and C10 propagate the myth of unicorns being attracted to virgins, which, while a part of medieval folklore, is presented without the crucial context that unicorns are not real.

As seen in Figure \ref{fig:generated-tree-with-reward}, the highest-reward nodes in the first layer (A3 and A4) lead to the generation of better children (B6 and B8), which in turn produce high-quality grandchildren (C2 and C7). This illustrates how generating from partial responses with high rewards tends to yield children nodes with similarly high rewards.

\subsection{Token-Level Reward Analysis}\label{appendix:tokenrmanalysis}
\begin{figure}[h]
    \centering
    \includegraphics[width=0.45\linewidth]{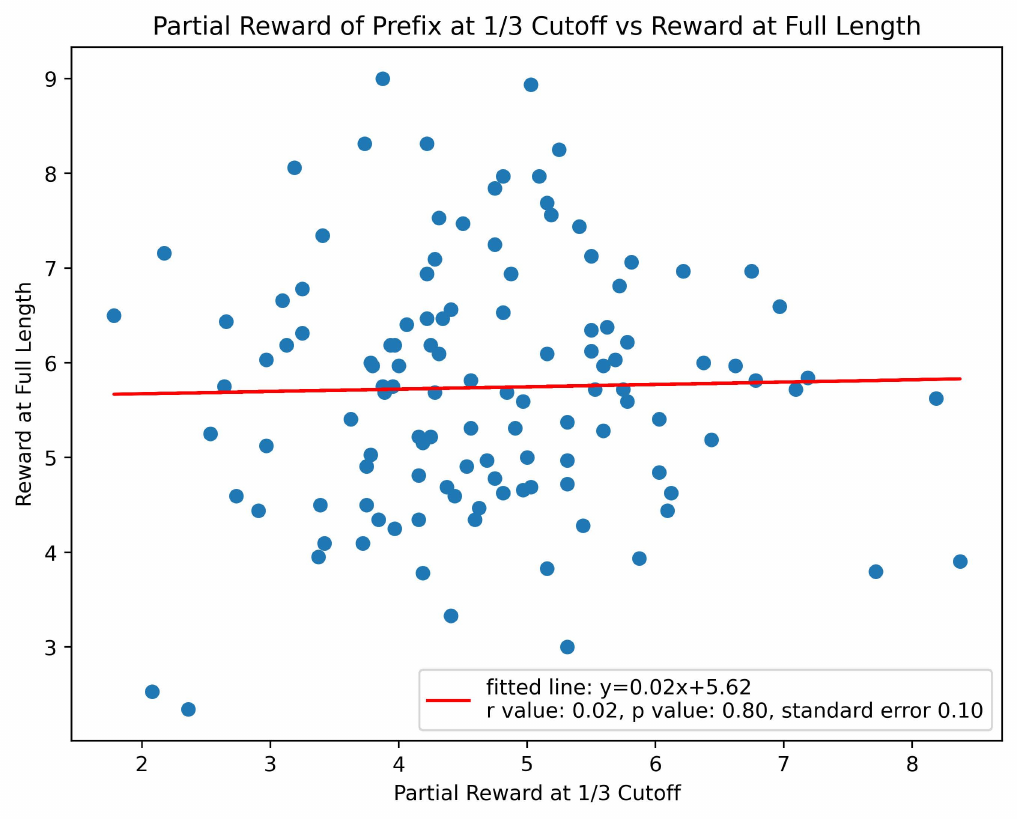}
    \includegraphics[width=0.45\linewidth]{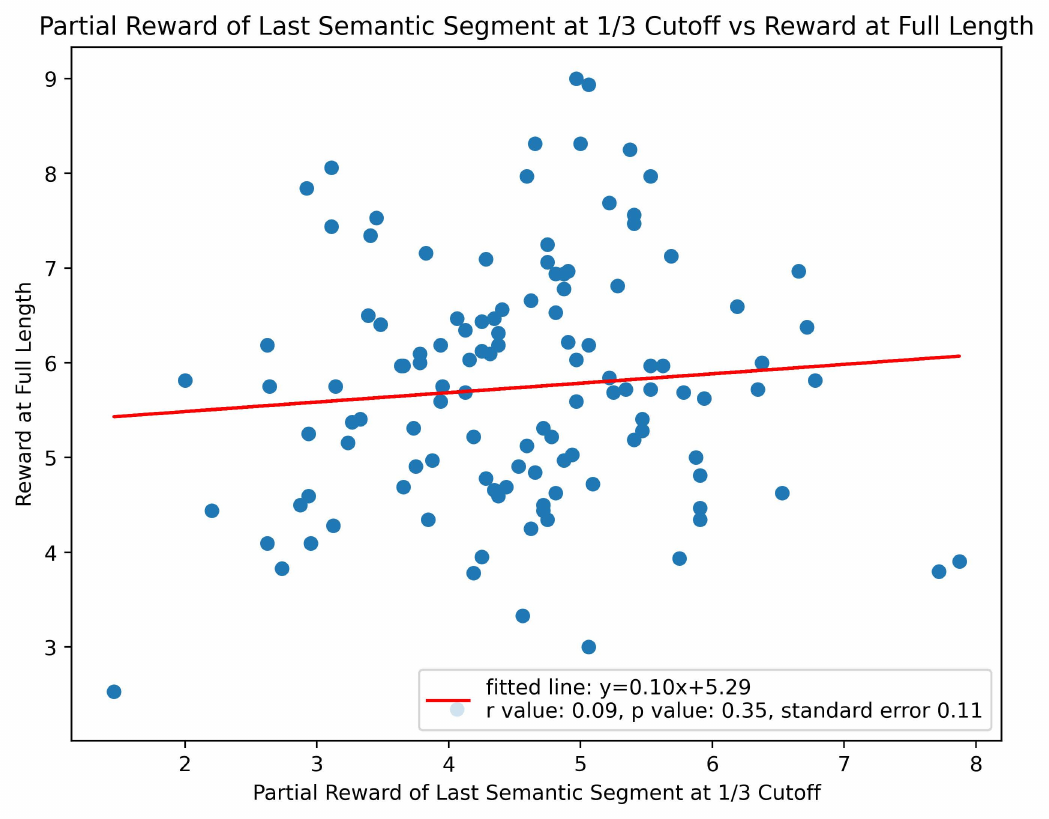}
    \caption{Partial rewards of prefixes of exactly 1/3 of the length, and prefixes before 1/3 that end with semantic complete segments against the reward of the full response with linear regressions.}
    \label{fig:reward_correlation}
\end{figure}

\begin{figure}[h]
    \centering
    \includegraphics[width=0.45\linewidth]{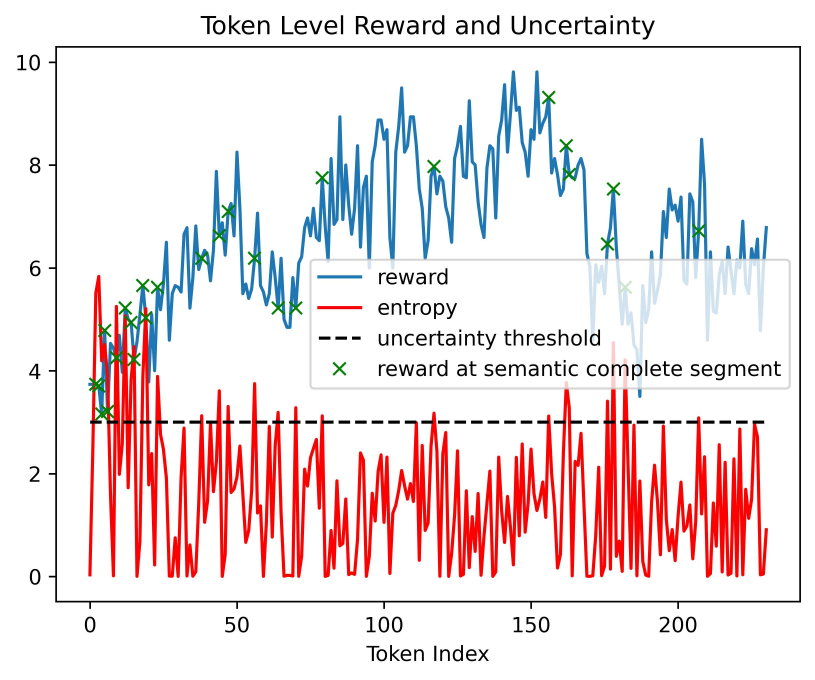}
    \includegraphics[width=0.45\linewidth]{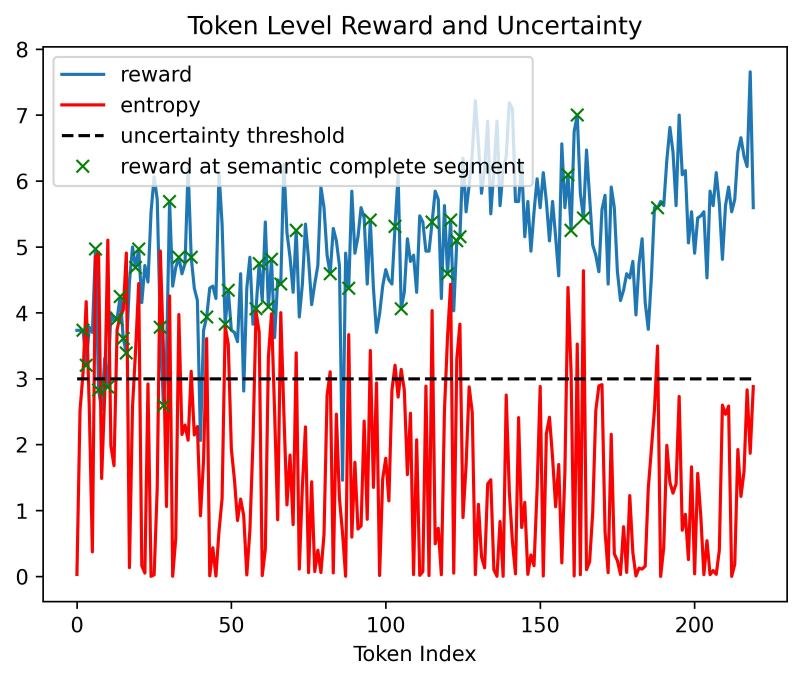}
    \caption{Token-level rewards at each token index for two responses generated by BoN with max new length 192, with the entropy along with the threshold. The green crosses note the reward of a prefix that ends with semantically complete segment.}
    \label{fig:token_level}
\end{figure}

In this section, we provide rationales to apply implicit reward from DPO instead of a trained reward model and analyze partial reward at token level, including using the concept of the semantically complete segment from Cascade Reward Sampling~(CARDS)\citep{li2024cascade}. We follow the setting in~\cite{li2024cascade}, use llama-7b-rm-float32\footnote{https://huggingface.co/argsearch/llama-7b-rm-float32} as the reward model, the entropy of LLM logits at each token as predictive uncertainty, and uncertainty threshold as 3, meaning that if a token has entropy greater than 3, we determine that it is at the end of a semantically complete segment. To verify the usability under our tree search setting. we then analyze different responses generated by BoN given one prompt from AlpacaFarm~\citep{alpacafarm}. In Figure~\ref{fig:reward_correlation}, partial rewards of prefixes of exactly 1/3 of the length, and prefixes before 1/3 that end with semantic complete segments, are plotted against the reward of the full response. Standard linear regressions are performed for both scatter plots. The reward and partial rewards are computed by the reward model on the responses generated by BoN with a max new length of 192. From the linear regression, we can see that though partial rewards of semantic complete prefixes have a slightly higher coefficient of correlation, the correlation is still very weak, and we can conclude that there is barely any correlation between partial rewards of prefixes and the rewards of full responses.  thus we conclude that the assumption in SBON~\citep{zhang2024accelerating} does not hold. In Figure~\ref{fig:token_level}, token-level rewards at each token index are plotted for two responses generated by BoN with a max new length of 192. Token-level rewards are computed by the reward model on all prefixes created from the first token until every token position. The entropy, which serves as a measure of predictive uncertainty, is also plotted, along with 3 being the threshold. The green crosses note the reward of a prefix that ends with the semantically complete segment. We can see that the token-level rewards fluctuate rapidly, and the partial rewards at the end of the semantically complete segment are not showing any better trend either. Thus, we demonstrate that the reward model is not capable of generating scores on short prefixes that align well with the reward for the whole response, as it is only trained on complete sequences.

\section{Limitations}
While TreeBoN achieves robust improvements, it greatly relies on the high quality of the reward model on incomplete responses to accelerate the inference without losing performance by iterative expansion and pruning, which is also key to SBoN~\citep{zhang2024accelerating}. Though implicit reward from the DPO model provides a candidate solution for the token-level reward guidance, it can only compare responses with the same length. Also, the poorly trained DPO model and its SFT checkpoints would fail to provide good partial rewards. Therefore, the accurate reward modeling of partial responses is still an open question. Reinforcement learning may provide better solutions for partial reward modeling but suffers from the difficulty of training. 

\section{Computing Requirement}
All experiments can be performed on a single NVIDIA H100. Depending on the specific tree configurations, one run could take from 1 hour to 24 hours.
\end{document}